\pdfoutput=1
\documentclass[11pt, letterpaper, logo, onecolumn, copyright, numbering]{paper_improved}

\usepackage[utf8]{inputenc} 
\usepackage[T1]{fontenc}    
\usepackage[most]{tcolorbox}

\usepackage{hyperref}       
\usepackage{url}            
\usepackage{booktabs}       
\usepackage{amsfonts}       
\usepackage{nicefrac}       
\usepackage{xcolor}         
\usepackage{graphicx}

\usepackage{natbib}
\usepackage{verbatim}
\usepackage{inconsolata}
\usepackage{tabularx}
\usepackage{float}
\usepackage{multirow}
\usepackage{listings}
\usepackage{xcolor}
\usepackage{svg}
\usepackage[T1]{fontenc}
\usepackage{wrapfig}

\usepackage{times}
\usepackage{url}
\usepackage{wrapfig,lipsum}
\usepackage{graphicx}
\usepackage{color}
\usepackage{colortbl}
\usepackage{booktabs} 
\usepackage{multirow}
\usepackage{array}
\usepackage[linesnumbered,ruled,vlined]{algorithm2e}
\usepackage{amsfonts}       
\usepackage{xcolor}         
\usepackage{float}
\usepackage{caption}
\usepackage[table]{xcolor}
\newcolumntype{C}[1]{>{\centering\arraybackslash}p{#1}}

\usepackage{float}
\usepackage{ragged2e}  
\usepackage{subcaption}
\usepackage{tabularx}
\usepackage[most]{tcolorbox}
\usepackage{listings}
\usepackage{tcolorbox}

\newtcolorbox[%
    auto counter,          
    number within=section, 
    list inside=promptlist 
]{promptbox}[1][]{
  colback=blue!5!white,
  colframe=blue!75!black,
  title=Decision Rules,
  fonttitle=\bfseries,
  boxrule=0.8mm,
  arc=2mm,
  #1 
}

\definecolor{lightgray}{rgb}{0.95, 0.95, 0.95}
\definecolor{darkgray}{rgb}{0.4, 0.4, 0.4}
\definecolor{backcolour}{rgb}{0.95,0.95,0.92}
\definecolor{myblue}{rgb}{0.2, 0.4, 0.8} 
\definecolor{mygreen}{rgb}{0.2, 0.6, 0.2} 

\usepackage{amsmath,amssymb,amsfonts}
\usepackage{soul}
\usepackage{url}
\usepackage[utf8]{inputenc}
\usepackage{caption}
\usepackage{graphicx}
\usepackage{xcolor}

\usepackage{amsthm}
\usepackage{booktabs}
\usepackage{latexsym}
\usepackage{graphicx} 
\usepackage{microtype}
\usepackage[switch]{lineno}
\usepackage{alltt}

\definecolor{forestgreen}{rgb}{0.13, 0.55, 0.13}

\tcbset{
    aibox/.style={
        colback=white, 
        colframe=blue!75!black, 
        colbacktitle=blue!85!black, 
        coltitle=white, 
        fonttitle=\bfseries,
        enhanced,  
        drop shadow=black!50!white,  
        boxrule=1pt,  
        boxsep=10pt,  
        left=10pt,  
        right=10pt,  
        top=6pt,  
        bottom=6pt,  
        title code={\path[tcb fill frame] ([xshift=-10pt]frame.west) -- (frame.north west) -- (frame.north east) -- ([xshift=10pt]frame.east) -- cycle;},  
        attach boxed title to top left={xshift=10pt, yshift*=-\tcboxedtitleheight/2},
        boxed title style={boxrule=0pt, frame code={}}  
    }
}

\newtcolorbox{AIbox}[2][]{aibox, title=#2, #1}

\let\cite\citep

\title{\textsc{HSCodeComp}: A Realistic and Expert-level Benchmark for Deep Search Agents in Hierarchical Rule Application}

\author[*,1]{Yiqian Yang$^{\dagger}$, Tian Lan$^{\dagger}$, Qianghuai Jia{$^*$}, Li Zhu, Hui Jiang, Hang Zhu, Longyue Wang, Weihua Luo, Kaifu Zhang\\ \bf Alibaba International Digital Commerce \\
\vspace{1mm} $*$ Corresponding Author: Qianghuai Jia (qianghuai.jqh@alibaba-inc.com)\\
\vspace{1mm} $\dagger$ Equal Contribution: Yiqian Yang, Tian Lan\\
}

\begin{abstract}
{\bf \large Abstract}\vspace{1mm}

Effective deep search agents must not only access open-domain and domain-specific knowledge but also apply complex rules—such as legal clauses, medical manuals and tariff rules.
These rules often feature vague boundaries and implicit logic relationships, making precise application challenging for agents.
However, this critical capability is largely overlooked by current agent benchmarks.
To fill this gap, we introduce \textsc{HSCodeComp}, the first realistic, expert-level e-commerce benchmark designed to evaluate deep search agents in hierarchical rule application. 
In this task, the deep reasoning process of agents is guided by these rules to predict 10-digit Harmonized System Code (HSCode) of products with noisy but realistic descriptions.
These codes, established by the World Customs Organization, are vital for global supply chain efficiency. 
Built from real-world data collected from large-scale e-commerce platforms, our proposed \textsc{HSCodeComp} comprises 632 product entries spanning diverse product categories, with these HSCodes annotated by several human experts.
Extensive experimental results on several state-of-the-art LLMs, open-source, and closed-source agents reveal a huge performance gap: best agent achieves only 46.8\% 10-digit accuracy, far below human experts at 95.0\%. 
Besides, detailed analysis demonstrates the challenges of hierarchical rule application, and test-time scaling fails to improve performance further.
\\
\\
\vspace{3mm}
\makebox[1pt][l]{\parbox{\textwidth}{\raggedright\begin{tabular}{@{} l l @{}} \raisebox{-0.5em}{\includegraphics[height=1.6em]{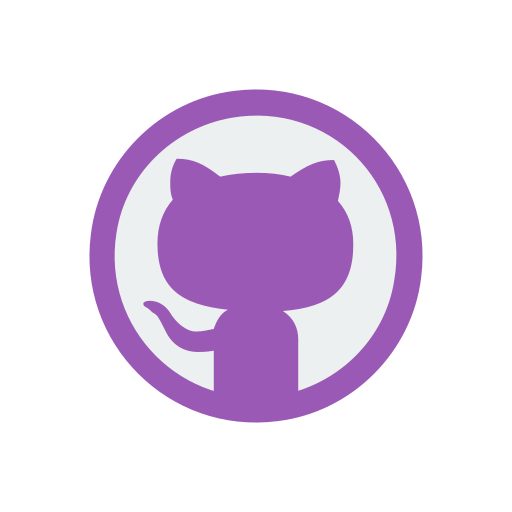}} & \small\url{https://github.com/AIDC-AI/Marco-Search-Agent}\\ \raisebox{-0.3em}{\includegraphics[height=1.4em]{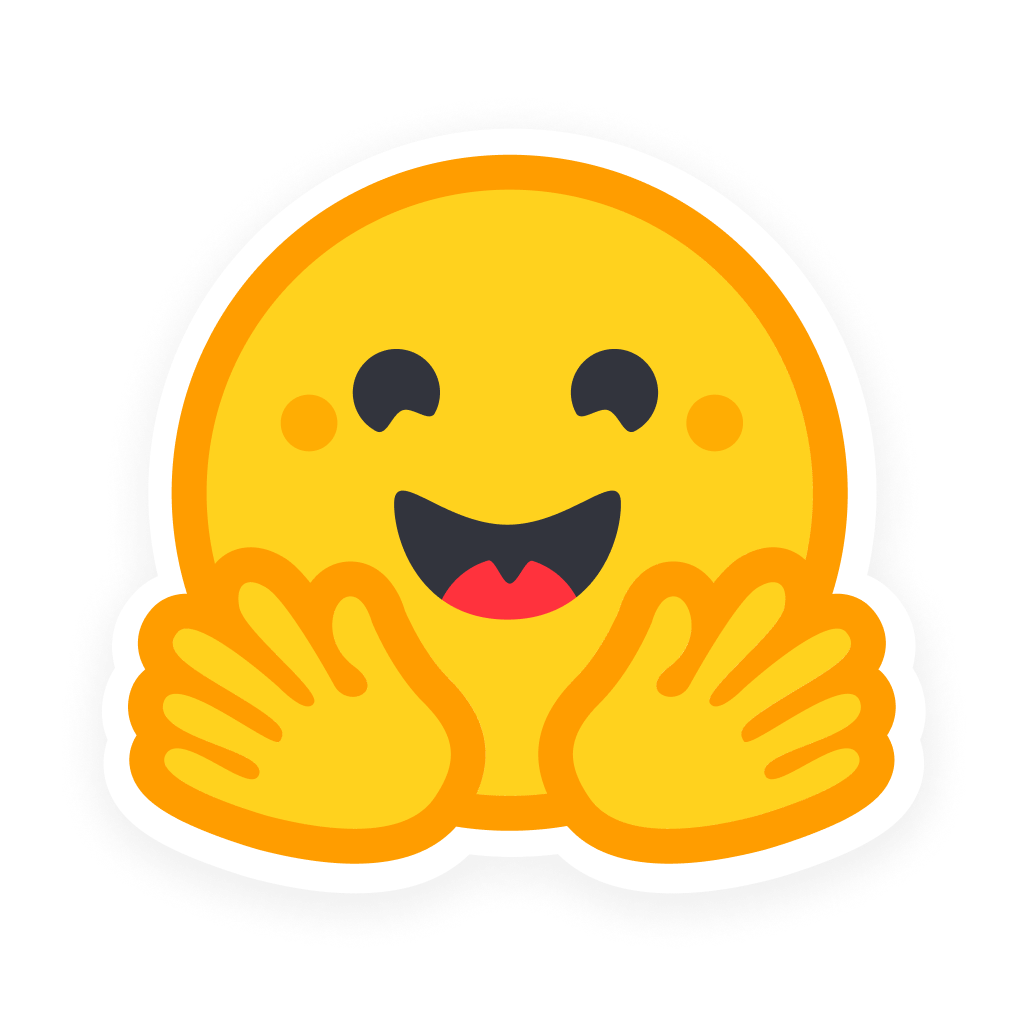}} &\small\url{https://huggingface.co/datasets/AIDC-AI/HSCodeComp} \end{tabular}}}

\end{abstract}

\begin{document}

\maketitle

\begin{figure}[ht]
  \raisebox{-0.5\height}{\includegraphics[width=0.6\textwidth]{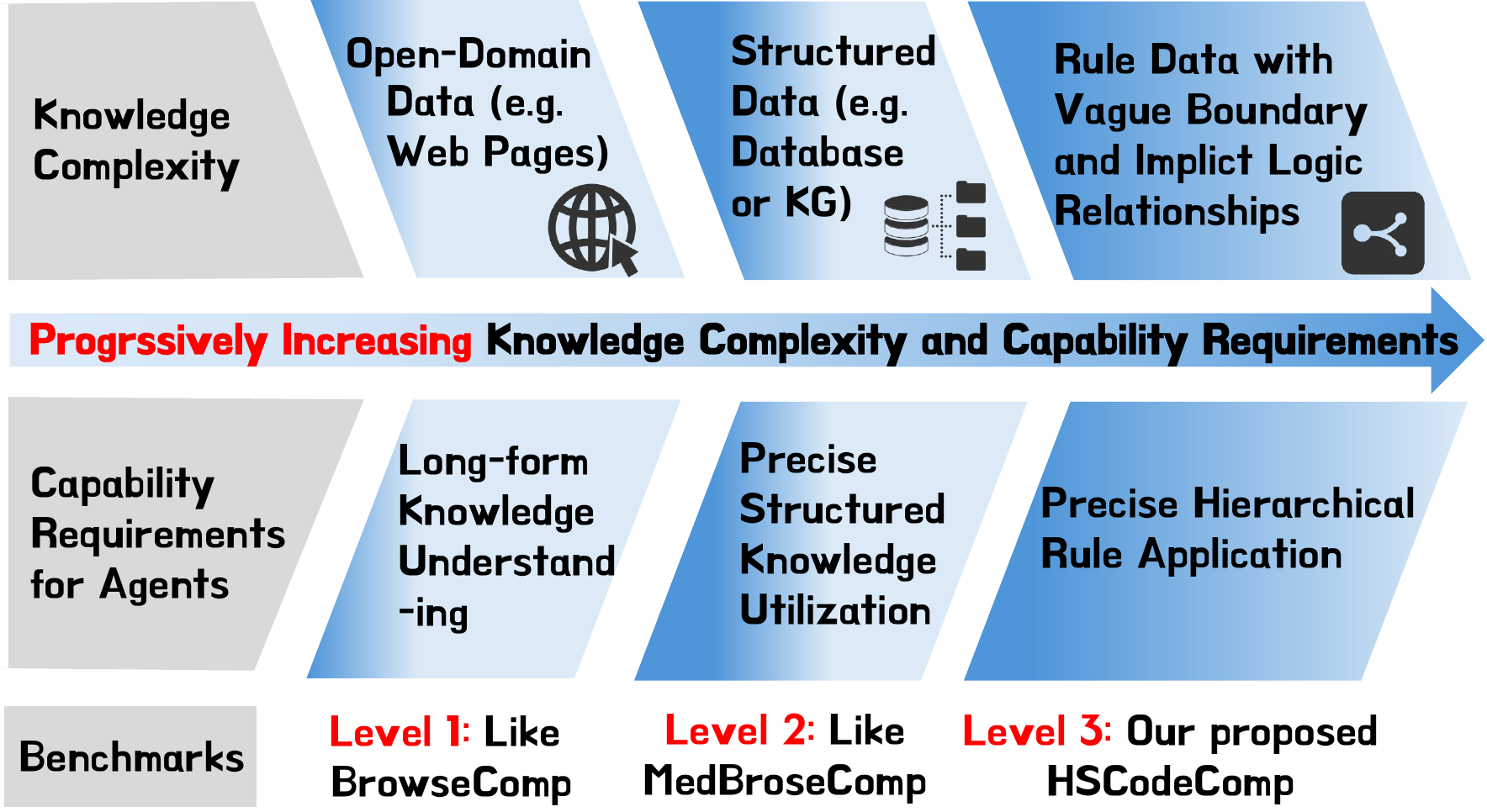}}
  \hfill
  \raisebox{-0.5\height}{\includegraphics[width=0.38\textwidth]{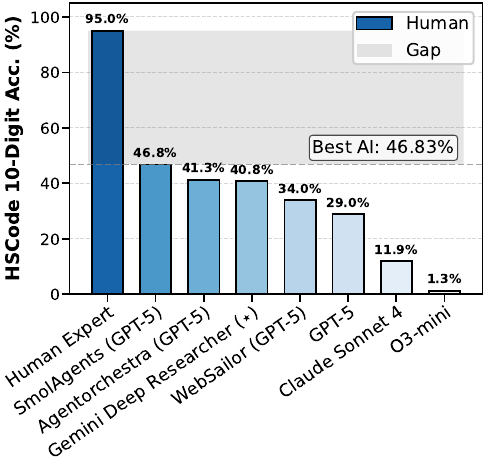}} 
  \vspace{-3pt}
  \caption{\textbf{Left:} Recent benchmarks reveal the increasing knowledge complexity and capability requirements for agents. \textbf{Right:} 10-digit HSCode accuracy of state-of-the-art baseline largely lags behind human experts (46.8\% $<$ 95.0\%), proving the challenges of hierarchical rule application. The closed-source agent ($\star$) is evaluated on the subset due to API unavailability.}
  \label{img:overview}
  \vspace{-15pt}
\end{figure}

\section{Introduction}\label{sec.introduction}

Deep search agents have demonstrated significant value in solving complex real-world problems, where robust external knowledge utilization constitutes a critical capability~\citep{wu2025webwalkerbenchmarkingllmsweb,tao2025webshaperagenticallydatasynthesizing,li2025websailornavigatingsuperhumanreasoning}. To evaluate this capability, numerous established benchmarks are proposed to assess agents in utilizing open-domain data (\textit{e.g.,} GAIA~\citep{mialon2023gaiabenchmarkgeneralai} and BrowseComp~\citep{wei2025browsecompsimplechallengingbenchmark}) and domain-specific data (\textit{e.g.,} WebMall~\citep{peeters2025webmallmultishopbenchmark}, FinSearchComp~\citep{hu2025finsearchcomprealisticexpertlevelevaluation} and MedBrowseComp~\citep{yu2025medresearcherr1expertlevelmedicaldeep}).

Beyond open-domain and domain-specific data, agents also need to effectively apply rules that encode human expert knowledge, particularly in scenarios like law, medical and e-commerce~\citep{legalagentbench2025,chen2025medbrowsecompbenchmarkingmedicaldeep,webshop,chollet2025arcagi2newchallengefrontier}. 
For instance, legal case adjudication require interpreting abstract legal provisions, and accurate e-commerce product classification in depends on tariff rules~\citep{grainger_wcj_2024}.
Previous works have defined rule application as using specific logical rules with supporting facts to derive conclusions~\citep{wang-etal-2024-symbolic,servantez-etal-2024-chain}. In contrast, we define it as a core capability for deep search agents, where human-written rules are systematically applied to guide complex reasoning and decision-making ~\citep{sadowski2025explainableruleapplicationstructured}.
Building on this observation, we categorize knowledge data for deep search agents into three levels (Figure~\ref{img:overview}, left), with increasing knowledge complexity:
(1) \textit{Level 1: Open-domain Data} - Tests understanding and deep reasoning abilities of agents on long-form web content. Established benchmarks include GAIA~\citep{mialon2023gaiabenchmarkgeneralai} and BrowseComp~\citep{wei2025browsecompsimplechallengingbenchmark};
(2) \textit{Level 2: Structured Data} - Assesses agents to precisely utilize structured data such as databases and knowledge graphs, as seen in domain-specific benchmarks like WebMall~\citep{peeters2025webmallmultishopbenchmark}, MedBrowseComp~\citep{chen2025medbrowsecompbenchmarkingmedicaldeep} and FinSearchComp~\citep{hu2025finsearchcomprealisticexpertlevelevaluation};
(3) \textit{Level 3: Rule Data} - Evaluates agents to apply complex and abstract rules~\cite{chollet2025arcagi2newchallengefrontier}. This level presents two key challenges: (a) making accurate decisions when rules contain vague natural language descriptions~\citep{sadowski2025explainableruleapplicationstructured}; and (b) reasoning about logical dependencies among rules, such as exception clauses and cross-category relationships~\citep{guha2023legalbench}.
Despite the importance of rule application in real-world scenarios, current agent benchmarks largely overlook its evaluation.


To fill this gap, we introduce \textsc{HSCodeComp} (short for the \textbf{H}armonized \textbf{S}ystem \textbf{Code} (HSCode) \textbf{Comp}etition), the first realistic, expert-level e-commerce benchmark designed to evaluate agents in predicting complete 10-digit Harmonized System Code (HSCode) of the product, using hierarchical rules (\textit{e.g.,} eWTP tariff rules\footnote{\url{https://www.ewtp.com/web/smart/hscode}}).
HSCodes organize products through a hierarchical structure spanning over 5,000 distinct codes across multiple classification levels, representing the global standard for classifying traded international goods, established by the World Customs Organization and implemented across more than 200 countries for customs clearance and tariff determination~\citep{grainger_wcj_2024,nath2025domainadaptivesmalllanguagemodels}.
Built from the data of the large-scale e-commerce platforms, our proposed \textsc{HSCodeComp} comprises 632 carefully curated product entries, encompassing 27 unique HS chapters and 32 distinct first-level categories. These HSCodes have been rigorously annotated by multiple e-commerce domain experts, ensuring that \textsc{HSCodeComp} is expert-level.
Accurately predicting the exact 10-digit HSCode presents significant challenges: agents must perform multi-hop hierarchical reasoning with complex tariff rules while processing noisy but realistic product descriptions that often contain abbreviations, language variations, or incomplete information.

Extensive experiments on the state-of-the-art baselines, including 14 advanced foundation models, 6 advanced open-source agent systems and 3 closed-source agent systems, demonstrate that HSCode prediction task remains a substantial challenge for current AI approaches.
As shown in the Figure~\ref{img:overview} (right), even the best-performing system (SmolAgent~\citep{smolagents} with GPT-5) achieves only 46.8\% accuracy, substantially below the 95.0\% accuracy attained by human experts. 
Further detailed analysis reveals that existing agent systems lack critical capabilities required for this complex hierarchical rule applications.
Notably, test-time scaling approach—which has proven effective in other reasoning tasks~\citep{guo2025rewardreasoningmodel,liu2025inferencetimescalinggeneralistreward}—fail to improve performance on \textsc{HSCodeComp}.
These observations demonstrate the challenging nature of our proposed \textsc{HSCodeComp}, highlighting the need for more effective designs of agent systems.
To facilitate future research, we will publicly release codes and the benchmark dataset of \textsc{HSCodeComp}.

\section{Related Works}
\label{sec:relatedworks}

\subsection{Previous Works in HSCode Prediction}
Previous works treat HSCode prediction as the e-commerce text classification task~\citep{grainger_wcj_2024}, using pre-trained BERT models~\citep{eblcs2024,ibm_hs_ensemble} or Large Language Models (LLMs) prompting~\citep{hussain2023auto_chatgpt}. However, these approaches fail to leverage domain-specific knowledge, especially the rules written by human experts~\citep{hussain2023auto_chatgpt,hts_benchmark}. 
Besides, existing HSCode benchmarks face two critical limitations~\citep{hts_benchmark,lee2024explainable,alma9922963629202321}: (1) they are typically constructed from publicly accessible customs rulings, suffering from data leakage; (2) they are not released.
In contrast, our released \textsc{HSCodeComp} is collected from large-scale online shopping platforms with noisy product descriptions, making it more challenging and realistic.

\subsection{Benchmarking Level 1 Knowledge Utilization}
Numerous benchmarks have been proposed to evaluate agent capabilities in understanding and deep reasoning over long-form open-domain web content~\citep{webgames,webvoyager,triviaqa,hle}.
For example, WebArena~\citep{webarena} provides realistic, self-hostable websites with standardized evaluation protocols to assess functional correctness.
WebShop~\citep{webshop} and ALFWorld~\citep{alfworld} evaluate long-horizon decision-making abilities of agents in web environments through tool interactions.
More recent deep search benchmarks, such as GAIA~\citep{gaia}, BrowseComp~\citep{wei2025browsecompsimplechallengingbenchmark}, WebWalkerQA~\citep{wu2025webwalkerbenchmarkingllmsweb} and BrowseComp-ZH~\citep{zhou2025browsecompzhbenchmarkingwebbrowsing}, demand advanced tool-usage and deep reasoning capabilities~\citep{zhang2025agentorchestrahierarchicalmultiagentframework,li2025websailornavigatingsuperhumanreasoning}.

\subsection{Benchmarking Level 2 Knowledge Utilization}

Recent works have focused on how agents utilize structured knowledge in domain-specific applications. Unlike open-domain data, domain-specific knowledge is typically organized into structured formats such as databases and knowledge graphs~\citep{crmarena2025,medresearcher_r1,chen2025xbenchtrackingagentsproductivity}, enabling more precise knowledge retrieval and utilization. To evaluate these capabilities, numerous deep search benchmarks have been proposed, including 
WebMall~\citep{webmall2025} and DeepShop~\citep{lyu2025deepshopbenchmarkdeepresearch} for e-commerce, LegalAgentBench for law~\citep{legalagentbench2025}, 
FinSearchComp for finance~\citep{hu2025finsearchcomprealisticexpertlevelevaluation},
DAgent for data analysis~\citep{xu2025dagentrelationaldatabasedrivendata}, CRMArena for CRM workflows~\citep{crmarena2025}, 
and MedBrowseComp for medicine~\citep{chen2025medbrowsecompbenchmarkingmedicaldeep}.

In summary, while there exists numerous evaluation benchmarks for assessing agent performance in open-domain or domain-specific scenarios, none evaluates the ability to apply Level 3 abstract rule-based knowledge.
To address this critical gap, we introduce a realistic and expert-level e-commerce benchmark \textsc{HSCodeComp}. Our benchmark presents significant challenges even for state-of-the-art closed-source and open-source agent systems.

\section{Task Formulation of HSCode Prediction}\label{task}

The HSCode prediction task is to assign a valid and unique 10-digit Harmonized System (HS) code to a given noisy but realistic product description.
The product HSCode plays the crucial role in e-commerce system. It is the global standard for classifying traded goods, essential for tariffs, customs, and trade governance.
The core challenge is to learn a mapping function, \textit{i.e.,} agents implemented by Large Language Models (LLMs) or Vision Language Models (VLMs), $f: \mathcal{X} \to \mathcal{Y}$.

\begin{figure*}[ht]
    \centering
    \includegraphics[width=1\linewidth]{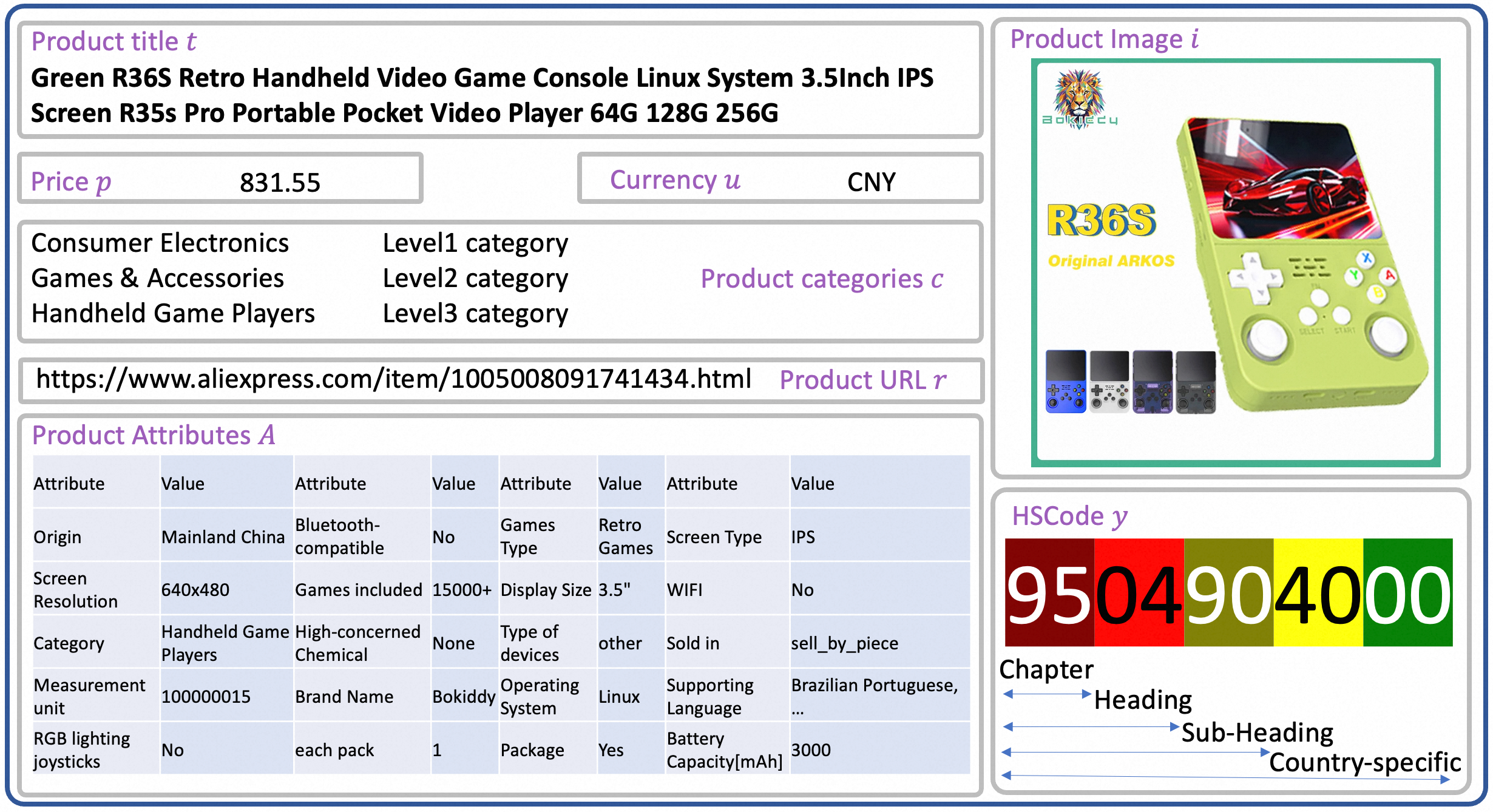}
    \caption{One example of a game console product in \textsc{HSCodeComp}.}
    \label{fig:example_figure_of_item_infomation}
    \vspace{-9pt}
\end{figure*}

\paragraph{Input:} Figure~\ref{fig:example_figure_of_item_infomation} shows that each product $x\in \mathcal{X}$ encompass rich records: $x=(t,A,c,i,p,u,r)$, where $t$ is the product title; $A=\{(k_j,v_j)\}_{j=1}^K$ represents a set of $K$ product attributes (\textit{e.g.,} material and package size); $c$ represents product categories defined by the e-commerce platform; $i$ is the product image; $p$ and $u$ are the price and currency; and $r$ is the webpage URL of the product.

\paragraph{Hierarchical Rule Utilization:} Accurate HSCode prediction requires effectively utilizing three types of e-commerce knowledge:
(1) \textbf{Hierarchical tariff rules} from official classification systems (\textit{e.g.,} eWTP), which organize product categories in a hierarchical structure.
As shown in Figure~\ref{fig:tariff_rules}, these rules contain complex implicit logic relationships, for example, the exception clause in tariff rules like \textit{excluding articles of HS heading 8539 ...} (highlighted with red boxes).
Besides, these rules often employ vague linguistic constraints (highlighted with blue boxes) that challenge existing AI agents;
(2) \textbf{Human-written decision rules} that specify how to correctly apply tariff rules. These rules provide high-level decision principles (see Figure~\ref{prompt_example} for an example with six key principles defined by domain experts); and 
(3) \textbf{Official customs rulings databases}, such as the U.S. Customs Rulings Online Search System (CROSS)\footnote{\url{https://rulings.cbp.gov/}}, which document historical HSCode classification decisions. As illustrated in Figure~\ref{fig:cross_example}, these databases contain complex information format requiring advanced reasoning capabilities.

\paragraph{Output:} The HSCode $y \in \mathcal{Y}$ is a single 10-digit numeric string $\mathcal{Y} \subseteq \{0,1,\ldots,9\}^{10}$.
The HSCode is hierarchical, where first 2-digit, 4-digit, and 6-digit represents the HS chapter, heading and sub-heading of tariff classification of products, respectively, and last 4 digits (from 6 to 10) are country-specific codes. 
In summary, this 10-digit HSCode follows a valid path in the official HS taxonomy.

\section{\textsc{HSCodeComp} Construction and Evaluation Metrics}
\label{sec.benchmark}


\subsection{Benchmark Construction}
\label{sec:benchmark-construction}

We design a rigorous pipeline, ensuring the dataset is diverse, realistic and expert-level: (1) Data Collection and Diversity Control; (2) Human Expert Annotation; and (3) Human Expert Validation.

\paragraph{Data Collection and Diversity Control.}
Products in our proposed \textsc{HSCodeComp} is sourced from a large-scale global e-commerce platform.
These product profiles include the noisy information, ensuring that task instances reflect the real-world challenges. 
Besides, we also balance the data category distribution to prevent the potential topical skew. Specifically, we apply a pre-processing step: a semantic redundancy filter discards the products ($x$) sharing identical categories ($c$) and 10-digit HSCode ($y$) with existing product instances. This ensures that \textsc{HSCodeComp} is not dominated by common and easy-to-classify products.

\begin{figure}[ht]
    \centering
    \includegraphics[width=1\linewidth]{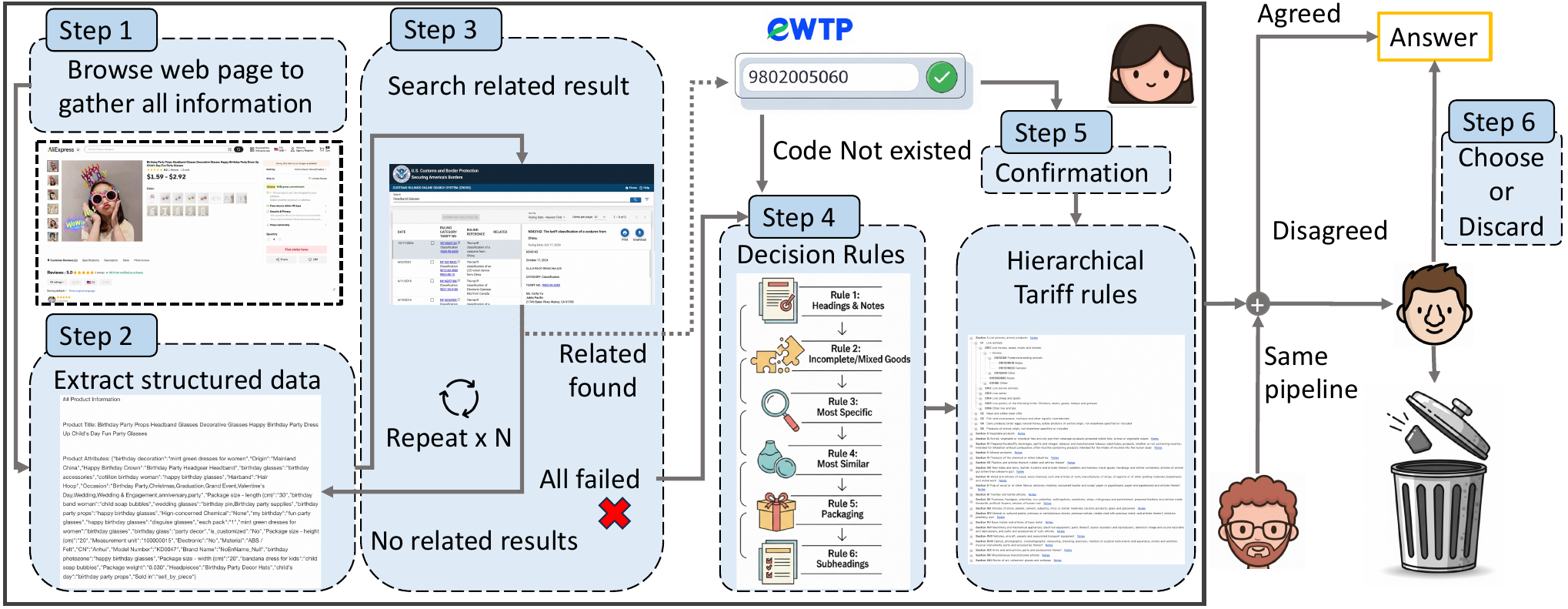}
    \caption{The pipeline for human experts to annotate the HSCodes, including two human experts for HSCode annotation (Step 1 to 5) and one additional expert for quality validation (Step 6).}
    \label{fig:original_label_sop}
\end{figure}

\paragraph{Human Expert Annotation.}
To ensure the quality of \textsc{HSCodeComp}, we engage human experts specialized in HSCode classification to annotate the HSCode $(y)$ for each product profile ($x$). 
As shown in Figure~\ref{fig:original_label_sop} (left), the annotation process follows a five-step pipeline:
(1) Two experts gather comprehensive information from the product webpage (Step 1);
(2) They extract the core structured features of products (Step 2);
(3) Experts search the official customs ruling databases (CROSS) for related cases. If a related case is found (very rare during the annotation of \textsc{HSCodeComp}), the corresponding HSCode is then validated on eWTP system, followed by the minor revision. Otherwise, they refine their search queries or revisit Step 2 to adjust the extracted features (Step 3);
(4) For products without any related cases, experts execute human-written hierarchical decision rules to apply tariff rules, and determine the appropriate HSCode (Step 4);
(5) Finally, experts verify the final identified HSCodes on the eWTP website to ensure its validity (Step 5).

\paragraph{Human Expert Validation.}
As shown in the Figure~\ref{fig:original_label_sop} (Step 6), when the HSCodes assigned by two experts match, the code is accepted. When they disagree, a senior tariff expert reviews both annotations to determine the correct HSCode. If neither annotation is valid, the instance is excluded from the dataset.
Finally, we collect 632 products with their human-annotated corresponding HSCodes, spanning 32 first-level product category defined by a large-scale ecommerce platform and 27 HS chapters defined by eWTP\footnote{Please refer to Figure~\ref{fig:lv1_hs2} for more details about \textsc{HSCodeComp} statistics.}.
Furthermore, to verify the reliability of our process, we conducted an additional quality review. A fourth senior expert, not involved in the initial annotation, re-annotated a random 10\% sample of the dataset. This review shows only a 2\% disagreement rate, confirming the effectiveness and consistency of our dataset construction pipeline.

\subsection{Evaluation Metric}
\label{sec:auto-eval}

We conduct the exact match to compare the normalized HSCodes extracted from the final output against human-annotated ground truth. 
Our primary evaluation metric is the 10-digit HSCode accuracy, which measures whether the predicted code exactly matches the reference 10-digit code. 
Additionally, we also report accuracies at 2-digit, 4-digit, 6-digit and 8-digit levels to provide more comprehensive insights into the performance across different granularities.

\begin{table}[htb]
\centering
\small
\begin{tabular}{@{}p{3cm}C{1cm}C{1.2cm}C{1.2cm}C{1.2cm}C{1.2cm}C{1.5cm}@{}}
\toprule
\multicolumn{1}{c}{\multirow{2}{*}{\textbf{Baselines}}} & \multicolumn{1}{c}{\multirow{2}{*}{\textbf{\begin{tabular}{@{}c@{}}Model\\Type\end{tabular}}}} & \multicolumn{5}{c}{\textbf{HSCode Prediction Accuracy}} \\ \cmidrule(l){3-7}
\multicolumn{1}{c}{} & \multicolumn{1}{c}{} & \textbf{2-digit} & \textbf{4-digit} & \textbf{6-digit} & \textbf{8-digit} & \textbf{10-digit} \\ \midrule
\multicolumn{7}{c}{\textbf{LLM/VLM-Only}}  \\
\midrule
GPT-5                                 & VLM & 82.12 & 70.89 & 59.97 & 41.46 & 29.27 \\
Gemini-2.5-PRO                        & VLM & \textbf{82.28} & 71.04 & 59.02 & 40.51 & 24.21 \\
GPT-4o                                & VLM & 78.01 & 64.08 & 48.10 & 29.75 & 18.51 \\
Claude Sonnet 4                       & VLM & 78.80 & 64.08 & 45.25 & 22.63 & 11.23 \\ \midrule
GPT-5                                 & LLM & 82.59 & 69.78 & 56.33 & 40.98 & 28.96 \\
Gemini-2.5-PRO                        & LLM & 80.54 & 69.94 & 58.54 & 40.35 & 23.42 \\
GPT-4o                                & LLM & 75.47 & 61.55 & 45.73 & 30.06 & 18.35 \\
Claude Sonnet 4                       & LLM & 78.80 & 62.97 & 44.94 & 23.58 & 11.87 \\\midrule
Kimi-K2                               & LLM & 78.01 & 62.03 & 44.15 & 24.53 & 12.18 \\
DeepSeek-R1                           & LLM & 77.22 & 61.71 & 38.45 & 16.77 & 6.65  \\
DeepSeek-V3                           & LLM & 77.06 & 54.43 & 32.28 & 17.25 & 6.49  \\
Qwen-MAX                              & LLM & 71.52 & 48.58 & 24.21 & 11.23 & 3.80  \\
Qwen3-235B-A22B                       & LLM & 66.93 & 49.53 & 24.53 & 6.01  & 1.74  \\
O3-mini                               & LLM & 77.22 & 56.17 & 24.53 & 6.65  & 1.27  \\
Qwen3-32B                             & LLM & 64.40 & 29.27 & 8.07  & 1.27  & 0.32  \\
QWQ-32B                               & LLM & 66.77 & 29.11 & 4.43  & 1.42  & 0.16  \\
Qwen2.5-72B                           & LLM & 20.73 & 12.34 & 3.80  & 1.42  & 0.16  \\
Nemotron-32B                          & LLM & 43.51 & 5.70  & 0.16  & 0.00  & 0.00  \\ \midrule
\multicolumn{7}{c}{\textbf{Open-source Agent System (GPT-5 Backbone)}}  \\ 
\midrule
SmolAgents                            & VLM & 82.06 & \textbf{72.06} & \textbf{62.38} & \textbf{52.38} & \textbf{46.83} \\
SmolAgents                            & LLM & \textbf{82.28} & 70.89 & 59.81 & 49.05 & 42.72 \\
Aworld                                & LLM & \textbf{82.28} & 70.41 & 59.18 & 48.58 & 41.30  \\
Agentorchestra                        & LLM & 82.12 & 70.73 & 60.44 & 47.78 & 41.30  \\
OWL                                   & LLM & 72.63 & 61.87 & 51.58 & 41.77 & 37.34 \\
WebSailor                             & LLM  & 81.64 & 70.56 & 57.27 & 43.98 & 35.44\\
Cognitive Kernel                      & LLM & 80.06 & 69.15 & 54.59 & 40.03 & 26.42 \\ 
\bottomrule
\end{tabular}
\caption{The complete results of state-of-the-art baselines in our proposed \textsc{HSCodeComp}.}
\label{tab.Performance}
\end{table}

\section{Experiments}

\subsection{Experimental Setup}
\label{sec:exp-setup}

We evaluate three kinds of advanced approaches on \textsc{HSCodeComp}:
\paragraph{LLMs/VLMs (no tools):} We test 14 foundation models (GPT-5, Gemini-2.5-PRO, GPT-4o, Kimi-K2~\citep{kimiteam2025kimik2openagentic}, Claude Sonnet~4, DeepSeek series~\citep{deepseekai2025deepseekr1incentivizingreasoningcapability}, Qwen variants~\citep{yang2025qwen3technicalreport}, O3-mini, Nemotron-32B~\citep{bercovich2025llamanemotronefficientreasoningmodels}) for HSCode prediction using only internal knowledge. For VLMs (GPT-4o, GPT-5, Claude Sonnet 4, Gemini 2.5 Pro), we provide product images to assess the impact of the visual information.


\paragraph{Open-source Agent Systems:} We evaluate six open-source frameworks (SmolAgents~\citep{smolagents}, Aworld~\citep{yu2025aworldorchestratingtrainingrecipe}, Agentorchestra~\citep{zhang2025agentorchestrahierarchicalmultiagentframework}, OWL~\citep{hu2025owloptimizedworkforcelearning}, WebSailor~\citep{li2025websailornavigatingsuperhumanreasoning} and Cognitive Kernel~\citep{fang2025cognitivekernelproframeworkdeep}) using GPT-5 as the default backbone. SmolAgent is enhanced with vision capabilities via product images, while Vision Language Models are incompatible with other agent frameworks. All frameworks utilize standardized tools including web search.


\paragraph{Closed-source Agent Systems:} We assess the performance of commercial systems Manus, Gemini Deep Research, and Grok DeepSearch. As these systems do not provide public APIs, we conduct manual evaluations on 49 representative examples from the \textsc{HSCodeComp} benchmark, following the evaluation protocol established in prior work~\citep{li2025websailornavigatingsuperhumanreasoning}.

All systems produce standardized outputs: a single HSCode in \texttt{\textbackslash boxed\{...\}} format. More implementation details appear in Appendix~\ref{appendix:implementation_detail}.

\subsection{Main Results\label{subsec.main_results}}

\begin{wraptable}[10]{r}{0.4\textwidth}
\small
\vspace{-10pt}
  \centering
\begin{tabular}{@{}p{4cm}C{1.6cm}@{}}
  \toprule
  \textbf{Agent Systems} & \textbf{10-digit}\\ \midrule
  Gemini Deep Researcher & 40.81 \\
  Manus & 30.61 \\
  Grok DeepSearch & 26.53 \\
  \midrule
  SmolAgents (GPT-5) & \textbf{42.86} \\
  Aworld (GPT-5) & \textbf{42.86} \\
  \bottomrule
  \end{tabular}
\caption{Comparison between closed-source and open-source agents.}
\label{tab:closed_source_exp}
\end{wraptable}

Table~\ref{tab.Performance} summarizes the performance of state-of-the-art LLM/VLM-Only models and open-source agents on \textsc{HSCodeComp}.
All approaches exhibit a consistent decline in accuracy as the HSCode length increases.
Notably, LLM/VLM-only baselines are much worse than agent systems due to their lack of domain-specific knowledge.
The best baseline, SmolAgent (GPT-5 VLM version), achieves only 46.83\% 10-digit accuracy, which remains substantially below the 95\% accuracy achieved by experienced human experts.
To ensure a fair comparison, we evaluate both closed-source and open-source agents on the same subset of \textsc{HSCodeComp}.
Table~\ref{tab:closed_source_exp} shows open-source agents outperform closed-source agents. 
Case studies reveal that closed-source agents suffer from the permature decisions and information misprocessing problems, as detailed in Section~\ref{subsec.failure_modes}.
In summary, the significant performance gap between human experts and the top-performing agent system underscores the challenges presented by \textsc{HSCodeComp}.
To better understand the factors affecting the performance, we conduct three ablation studies as below.
\begin{wraptable}[11]{r}{0.5\textwidth}
\small
\vspace{-10pt}
  \centering
    \begin{tabular}{@{}p{3.7cm}C{2cm}C{1.6cm}@{}}
    \toprule
    \textbf{Backbone Model} & \textbf{Model Type} & \textbf{10-digit}\\ \midrule
    SmolAgent \textbf{w/o DR} & VLM & \textbf{46.83}\  \\
    SmolAgent \textbf{w/ DR} & VLM & 43.83 $\boldsymbol{\downarrow}$ \\\midrule
    Aworld \textbf{w/o DR} & LLM & 41.30\  \\
    Aworld \textbf{w/ DR} & LLM & \textbf{42.95} $\boldsymbol{\uparrow}$ \\\midrule
    WebSailor \textbf{w/o DR} & LLM & \textbf{35.44} \\
    WebSailor \textbf{w/ DR} & LLM & 35.43 $\boldsymbol{\downarrow}$ \\
    \bottomrule
    \end{tabular}
    \caption{\label{tab:agent_DR_ablation}The ablation study on human-written \textbf{D}ecision \textbf{R}ules (\textbf{DR}). GPT-5 is the backbone.}
\end{wraptable}

\paragraph{Ablation Study on Hierarchical Decision Rules}
The hierarchical decision rules capture how human experts apply tariff rules. To assess whether agents can effectively leverage these rules, we conduct ablation experiments for following models: GPT-5, SmolAgent (GPT-5 VLM version), Aword (GPT-5 LLM version) and WebSailor (GPT-5 LLM version). 
As shown in Table~\ref{tab:agent_DR_ablation}, incorporating decision rules (w/ DR) decreases accuracy for both SmolAgent and WebSailor, while Aworld achieves only marginal gains. 
Therefore, we remove these decision rules for agents as the default setup.
These results indicate that current agent systems struggle at applying human-written decision rules, thereby limits their ability to utilize hierarchical tariff rules for HSCode prediction.

\begin{wraptable}[12]{r}{0.4\textwidth}
\small
\vspace{-12pt}
  \centering
    \begin{tabular}{@{}p{4.3cm}C{1.5cm}@{}}
  \toprule
  \textbf{Backbone Model} & \textbf{10-digit}\\ \midrule
  GPT-5 \textbf{w/o Image} & 42.72 $\ $ \\
  GPT-5 \textbf{w/ Image} & \textbf{46.83} $\boldsymbol{\uparrow}$ \\ \midrule
  Gemini-2.5-Pro \textbf{w/o Image} & \textbf{34.49} $\ $ \\
  Gemini-2.5-Pro \textbf{w/ Image} & 34.39 $\boldsymbol{\downarrow}$ \\ \midrule
  Claude 4 Sonnet \textbf{w/o Image} & 33.70 $\ $ \\
  Claude 4 Sonnet \textbf{w/ Image} & \textbf{34.65} $\boldsymbol{\uparrow}$ \\ 
  \midrule
  GPT-4o \textbf{w/o Image} & 22.03 $\ $\\
  GPT-4o \textbf{w/ Image} & \textbf{22.31} $\boldsymbol{\uparrow}$\\
  \bottomrule
  \end{tabular}
    \caption{The ablation study of the product images in SmolAgents.}
    \label{tab:agent_image_ablation}
\end{wraptable}
\paragraph{Multi-modal Information Is Helpful}
Table~\ref{tab.Performance} and Table~\ref{tab:agent_image_ablation} show that most baselines achieve consistent improvements when the product images can be accessed. 
Case studies in Appendix~\ref{appendix:multimodal_agent_ablation} show that understanding product images improves performance by capturing visual attributes—such as material and surface features—that are not present in the textual description but are critical for classification. These attributes align with the predefined rules, leading to performance gains.

\paragraph{Webpage Visits Decrease Agents Performance}
We augment SmolAgent (GPT-5 LLM version) with the capability to visit webpages, but this leading to 10-digit accuracy decrease from 42.72\% to 42.09\%. 
Our study reveals that a large amount of webpage content overwhelms the key information, misleading the agents, while this key information can be precisely extract by search engines in the snippets.
Therefore, we remove the webpage visit tool for all open-source agents as the default setup.

\section{Analysis on \textsc{HSCodeComp}}

We conduct several detailed analysis:
(1) Overthinking Decrease Open-Source Agent Performance (Section~\ref{subsec.overthinking});
(2) Effects of Backbones in agents (Section~\ref{subsec.open_source_agent});
(3) Failure Modes of Closed-Source and Open-Source Agents (Section~\ref{subsec.failure_modes});
(4) Per-category performance analysis (Section~\ref{subsec.distribution});
and (5) Effectiveness of Test-time Scaling  (Section~\ref{subsec.tts}).

\subsection{Overthinking Decrease Open-source Agent Performance\label{subsec.overthinking}}

\begin{wraptable}[9]{r}{0.6\textwidth}
\vspace{-15pt}
  \centering
  \begin{tabular}{@{}p{4cm}C{2.5cm}C{1.6cm}@{}}
  \toprule
  \textbf{Agent Systems} & \textbf{Think Depth} & \textbf{10-digit}\\ \midrule
  SmolAgent (GPT-5) & No-Think & 42.72 \\ 
  WebSailor (GPT-5) & No-Think & 40.82 \\
  WebSailor (GPT-5) & Medium-Think & 37.34 \\
  WebSailor (GPT-5) & Overthink & 35.44 \\
  \bottomrule
  \end{tabular}
    \caption{Agent performance with different think depth. GPT-5 LLM version is the backbone.
    \label{tab:overthink_ablation}}
\end{wraptable}

Table~\ref{tab.Performance} shows WebSailor underperforms SmolAgent despite using identical models and tools. Our case studies reveal that this occurs because WebSailor encourages excessive reasoning (\textbf{Overthink}), before tool-calling. Cases in Appendix~\ref{appendix:overthink} show that WebSailor often first conduct deep reasoning to predict the full 10-digit HSCode. The errors in reasoning significantly decreases the effectiveness of tool-calling. To prove this, we created two variants for WebSailor: (1) \textbf{No-Think}: direct tool calling without thinking; and (2) \textbf{Medium-Think}: medium-level reasoning depth before tool-calling. 
Medium-think denotes a moderate reasoning depth between No-think and Overthink. 
Table~\ref{tab:overthink_ablation} demonstrates that reducing reasoning depth improves accuracy, with No-Think nearly matching SmolAgent. 
This finding suggests that the primary factor contributing to performance variances among open-source agents is the reasoning depth defined in the task prompt.
For HSCode prediction, minimal reasoning with frequent tool calls outperforms extensive self-reasoning. When accurate information is available through calling tools, prioritizing tool utilization over reasoning yields better results for such complex domain-specific tasks.

\subsection{Effect of Backbones in Open-source Agent Systems\label{subsec.open_source_agent}}

\begin{wraptable}[9]{r}{0.35\textwidth}
\vspace{-12pt}
  \centering
    \begin{tabular}{@{}p{3cm}C{1.5cm}@{}}
    \toprule
    \textbf{Backbone Model} & \textbf{10-digit}\\ \midrule
    GPT-5& \textbf{42.72} \\
    Gemini-2.5-Pro& 34.49 \\
    Claude 4 Sonnet& 33.70 \\
    Qwen-MAX& 17.43 \\
    \bottomrule
    \end{tabular}
    \caption{The ablation study on the backbone models in SmolAgent.}
    \label{tab:agent_backbone_ablation}
\end{wraptable}
This subsection analyze the effects of the backbone LLM in the performance of agent systems. Specifically, we evaluate the performance of SmolAgents system implemented by four different backbone LLMs: GPT-5, Gemini-2.5-pro, Claude-4-Sonnet and Qwen-MAX. 
As demonstrated in Table~\ref{tab:agent_backbone_ablation}, different backbone LLMs yield markedly different results in the HSCode prediction task, and GPT-5 is the best backbone model. 
Therefore, we choose GPT-5 as the default setup for open-source agents. More results are provided in Appendix~\ref{appendix.smolagents_different_models}.

\subsection{Failure Modes of Closed-Source and Open-Source Agents}
\label{subsec.failure_modes}

We perform the qualitative and quantitative analysis for closed-source and open-source agents.

\paragraph{Qualitative analysis.} 
We identify six critical failure modes of open-source and closed-source agent systems in \textsc{HSCodeComp}:
(1) \textbf{Premature Decisions}: Agents commit to incorrect classification paths without collecting sufficient evidence (Table~\ref{tab:closed_source_case_2}, Grok DeepSearch);
(2) \textbf{Information Misprocessing}: Agents overlook or misinterpret key product details, indicating challenges with long-context processing (Table~\ref{tab:closed_source_case_1} and  Figure~\ref{fig:case_noise});
(3) \textbf{Unnecessary Self-Correction}: Agents sometimes predict correct HSCodes initially but revise them incorrectly through excessive critique (Table~\ref{tab:closed_source_case_1}, Gemini Deep Research);
(4) \textbf{Reasoning Hallucination}: Agents generate plausible but factually incorrect reasoning steps (Table~\ref{tab:closed_source_case_1}, Grok DeepSearch);
(5) \textbf{Wrong Rule Application}: Models frequently miss or misuse relevant tariff rules due to their ambiguous descriptions that confuse the reasoning process, resulting in incorrect classification decisions (Figure~\ref{fig:wrong_rule_application});
and (6) \textbf{Lack of Domain Knowledge}: Models exhibit errors due to insufficient domain-specific knowledge, such as misidentify silicone products as rubber instead of plastic (Figure~\ref{fig:lack_domain_knowledge}).
These limitations highlight that \textsc{HSCodeComp} remains challenging for advanced closed-source and open-source systems.

\begin{wrapfigure}[13]{r}{0.3\textwidth}
  \vspace{-10pt}
  \centering
  \includegraphics[width=\linewidth]{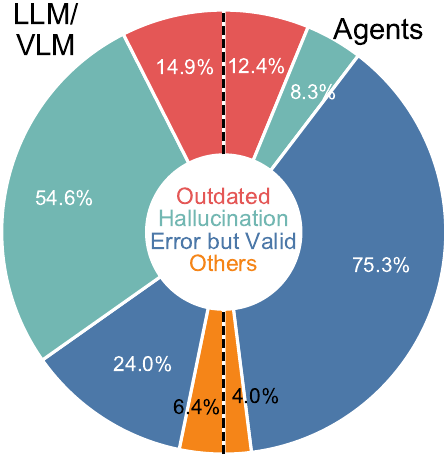}
  \caption{Failures analysis.}
  \label{fig:error_count_split}
\end{wrapfigure}

\paragraph{Quantitative analysis:} Figure~\ref{fig:error_count_split} present the distribution of four coarse-grained failures across both LLM/VLM-Only (left) and agents (right)\footnote{The average performance of LLM-only baselines and agent baselines are computed.}: (1) \textbf{Outdated}: Incorrect HSCodes due to changes in tariff rules over time;
(2) \textbf{Hallucination}: Invalid HSCodes that do not exist in the official coding system; 
(3) \textbf{Error but Valid}: HSCodes are valid and current, but differ from the ground-truth HSCodes;
and (4) \textbf{Others}: Other errors like wrong output formats, reaching maximum window and wrong tool-calling. Our analysis reveals that agents significantly reduce hallucination, outdated and other errors through effective tool utilization, compared with LLMs. Consequently, the predominant error type for agents is ``Error but Valid''.
Besides, Figure~\ref{fig:agent_llm_improvement} also quantifies the  improvements from GPT-5 to SmolAgent (GPT-5), demonstrating that the agent significantly reduces both outdated and hallucination errors.

\subsection{Per-category Performance Analysis\label{subsec.distribution}}
\begin{wrapfigure}[23]{r}{0.6\textwidth}
    \vspace{-15pt} 
    \centering
    \begin{subfigure}[b]{0.54\linewidth}
        \centering
        \includegraphics[width=\linewidth]{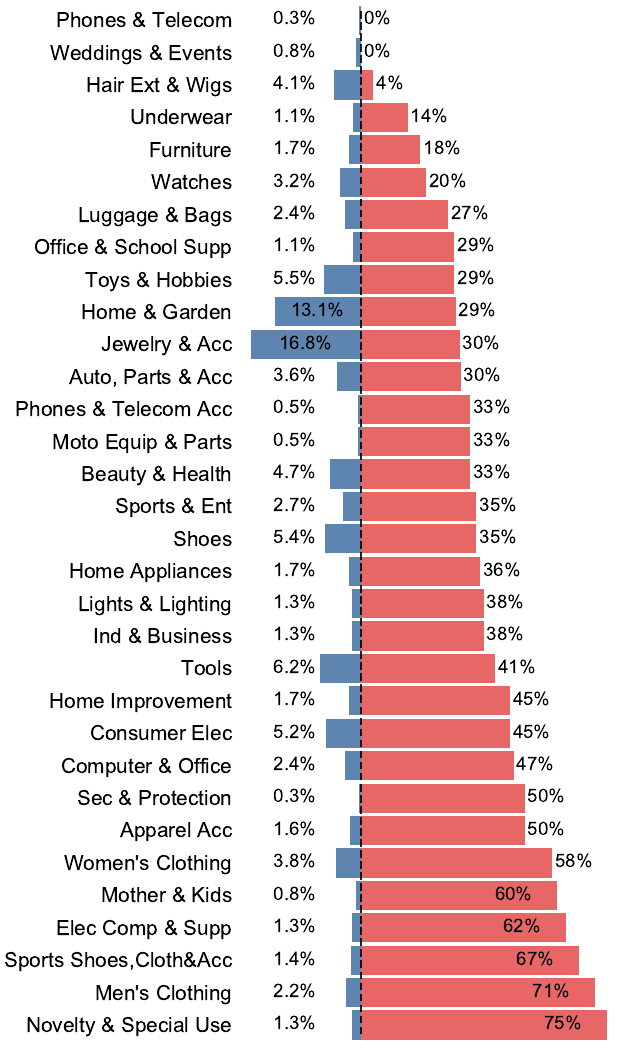}
        \label{fig:sub_wrong_ratio}
    \end{subfigure}
    \hfill 
    \begin{subfigure}[b]{0.43\linewidth}
        \centering
        \includegraphics[width=\linewidth]{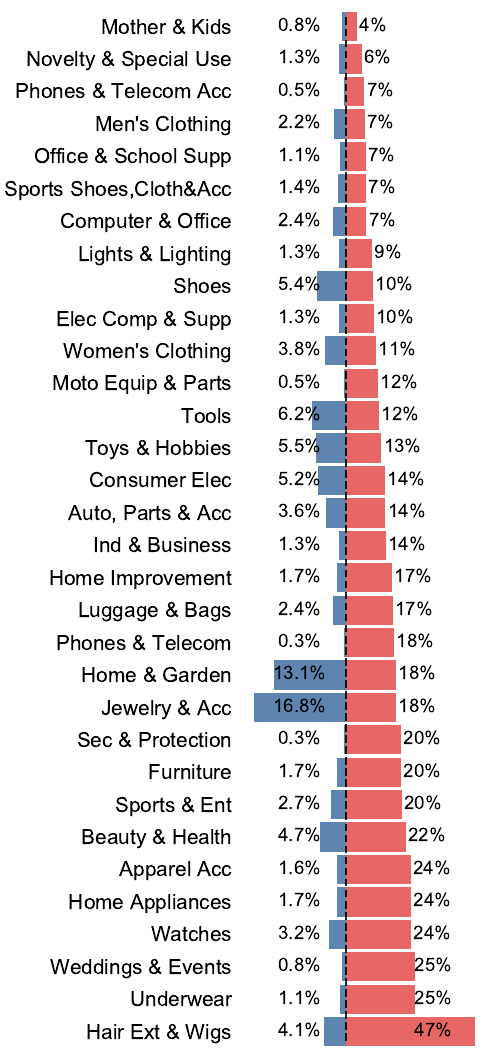}
        \label{fig:sub_mean_accuracy}
    \end{subfigure}
    \vspace{-15pt}
    \caption{Both figures show the category distribution on the left blue bars. \textbf{Left:} Challenging Product Distribution (CID). \textbf{Right:} Average Performance Distribution (APD).}
    \label{fig:combined_performance}
\end{wrapfigure}

We analyze two critical distributions across the 32 first-level product categories:
(1) \textbf{Challenging Product Distribution (CID)}: the distribution of products that all baseline methods failed to correctly predict;
(2) \textbf{Average Performance Distribution (APD)}: the distribution of average 10-digit accuracy across all baseline methods.
Figure~\ref{fig:combined_performance} reveals two key insights:
(1) The CID indicates that the most challenging products are concentrated in long-tail categories, such as \textit{Novelty \& Special Use} (1.3\%) and \textit{Men's Clothing} (2.2\%);
(2) The APD shows that average accuracy across most product categories remains below 25\%, with only \textit{Hair Extensions \& Wigs} achieving a relatively high accuracy of 47\%. Importantly, even for most frequent categories like \textit{Jewelry \& Accessories} (13.1\%), \textit{Home \& Garden} (16.8\%) and \textit{Tools} (6.2\%), the average performance stays below 18\%.
These findings underscore the challenges presented by \textsc{HSCodeComp}, highlighting the need for more robust and generalizable approaches to HSCode prediction.

\subsection{Test-time scaling Cannot Improve Performance Effectively}
\label{subsec.tts}

\begin{wrapfigure}[14]{r}{0.5\textwidth}
  \centering
  \includegraphics[width=\linewidth]{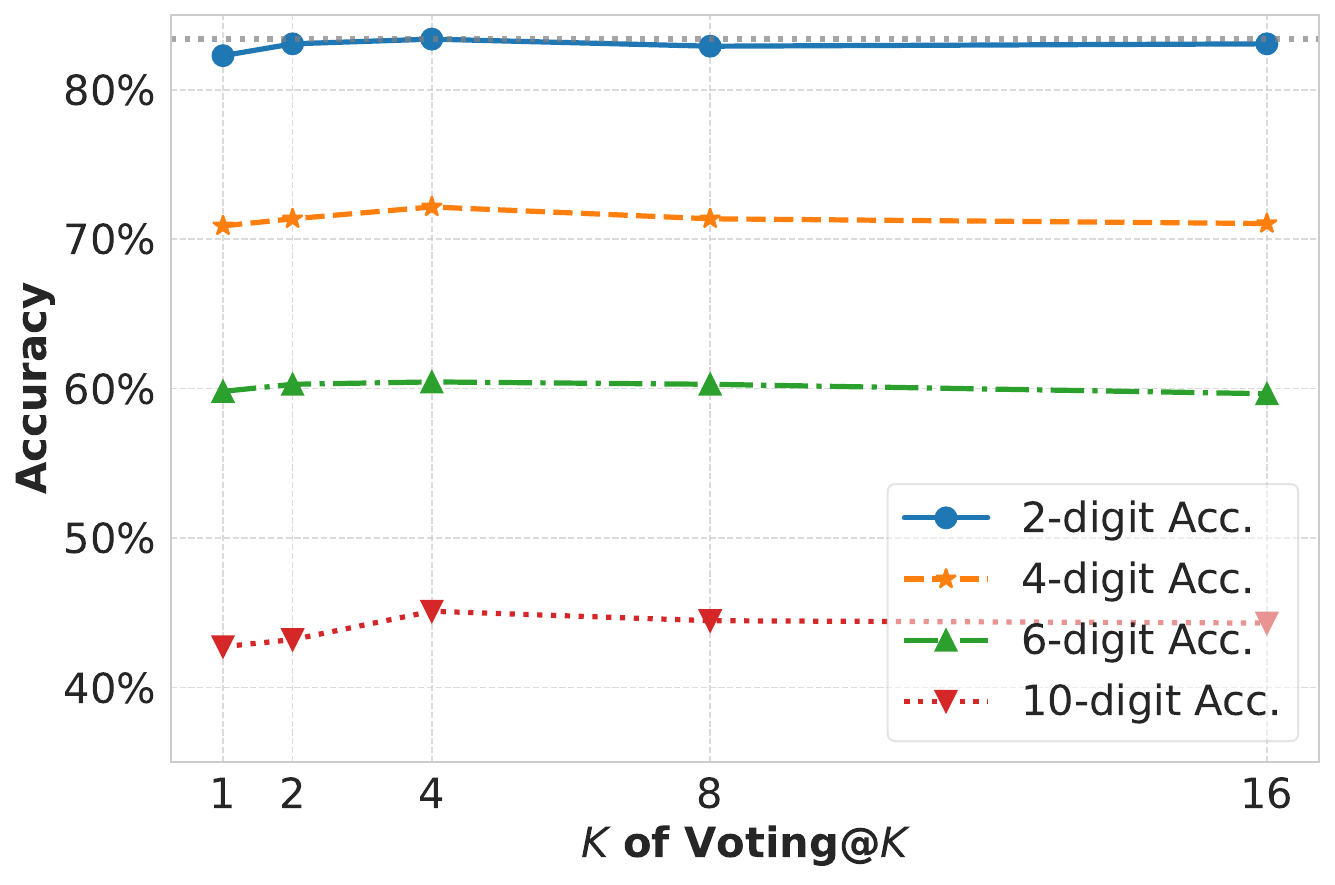}
  \vspace{-10pt}
  \caption{Majority voting experiments.}
  \label{fig:test_time_scaling}
\end{wrapfigure}

Test-time scaling (TTS) has demonstrated significant gains in complex reasoning tasks using more inference budget~\citep{liu2025inferencetimescalinggeneralistreward,guo2025rewardreasoningmodel,ma2025t2ievalr1reinforcementlearningdrivenreasoning}.
Given these successes, we investigate whether TTS can enhance performance on \textsc{HSCodeComp}. Specifically, we evaluate two established TTS strategies~\citep{liu2025inferencetimescalinggeneralistreward}: (1) \textbf{Majority Voting:} We implement majority voting across $K=\{1,2,4,8,16\}$ independent trials (Voting@$K$), using SmolAgent (GPT-5). Figure~\ref{fig:test_time_scaling} showns that increasing $K$ yields negligible performance improvement; (2) \textbf{Self-Reflection:} We integrate a self-reflection mechanism into SmolAgent (GPT-5), enabling the model to proactively evaluate and revise its reasoning and actions. However, this approach slightly decreases performance from 42.72\% to 42.57\%.
These results demonstrate a key limitation of standard TTS methods when applied to HSCode prediction, highlighting the need of more effective test-time scaling strategy for agents in hierarchical rule application.

\section{Conclusion}
\label{sec.conclusion}

We identified and addressed the critical gap in evaluating deep search agents in hierarchical rule applications. To address this gap, we introduced \textsc{HSCodeComp}, the first realistic and expert-level benchmark designed to assess agents for multi-hop reasoning with hierarchical tariff rules in e-commerce domain. Our extensive evaluation revealed a substantial performance gap between current state-of-the-art agents (46.8\%) and human experts (95.0\%), highlighting that hierarchical rule application remains a significant challenge for existing agent architectures. We will release the \textsc{HSCodeComp} to accelerate research in this crucial capability for real-world agent deployment.

\section{Ethics Statement}

This research adheres to strict ethical guidelines regarding data privacy and fair labor. The dataset is fully anonymized and contains no personally identifiable information. 
The hourly wage of our human annotators is over 34.6 USD, which
is much higher than average hourly wage 3.13 USD on Amazon Mechanical Turk~\citep{hara2017datadrivenanalysisworkersearnings}.
This remuneration structure was designed to provide a fair and competitive wage, acknowledging the expertise and effort required for this task and ensuring that contributors were rewarded appropriately for their work.

\section{Reproducibility statement}
We are committed to the principles of reproducible research. Accordingly, all necessary materials, including code, benchmark dataset and other related resources will be publicly released to promote the development of the deep search agents. 
For security and compliance reasons, the product URLs and Image have been removed from this version of our proposed \textsc{HSCodeComp} dataset.

\bibliography{references,custom}


\clearpage
\appendix
\section{The Use of Large Language Models (LLMs)}
In preparing this manuscript, Qwen-MAX and ChatGPT were used solely as a writing assistant to improve grammar and clarity. The LLMs was not used for generating code, concepts, or any part of the core research methodology.

\section{Dataset distribution}
\label{appendix.distribution}

\begin{figure}[H]
\vspace{-15pt}
  \centering
  \begin{subfigure}[t]{0.6\linewidth}
    \centering
    \includegraphics[height=0.4\textheight]{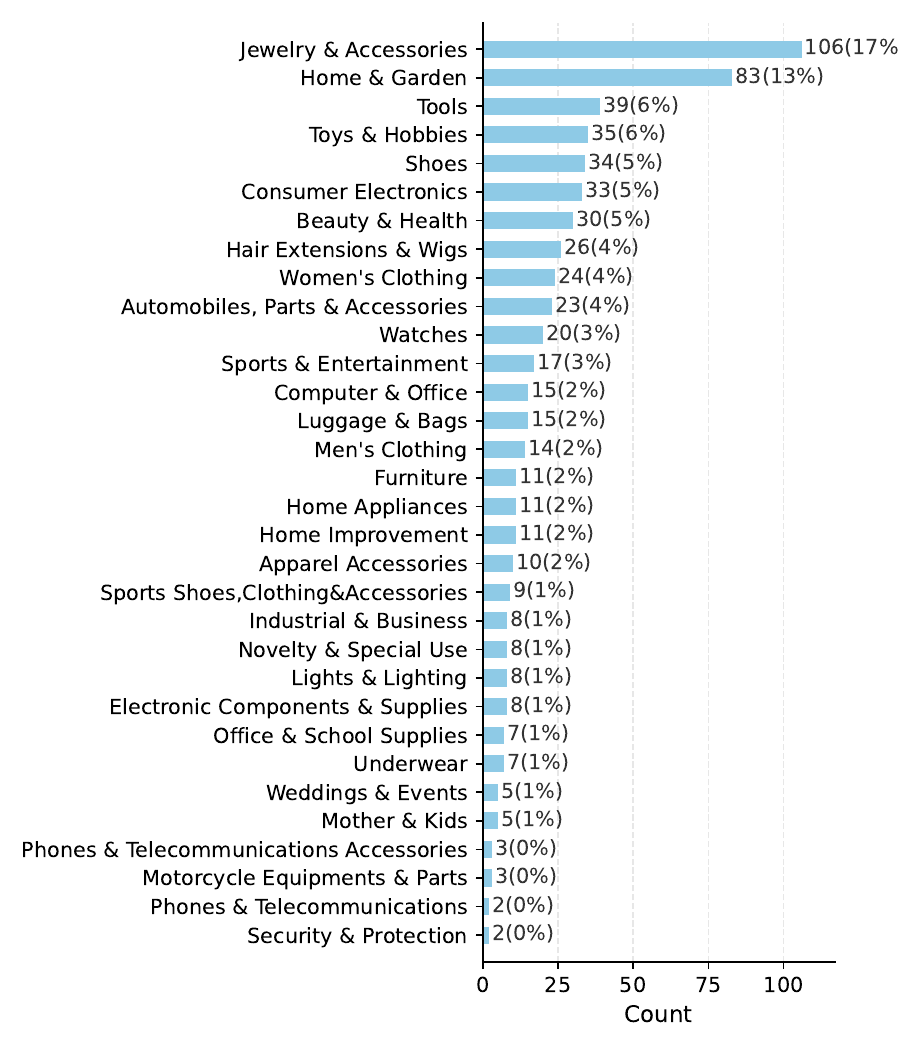}
    \caption{First-level product categories}
  \end{subfigure}\hfill
  \begin{subfigure}[t]{0.4\linewidth}
    \centering
    \includegraphics[height=0.395\textheight]{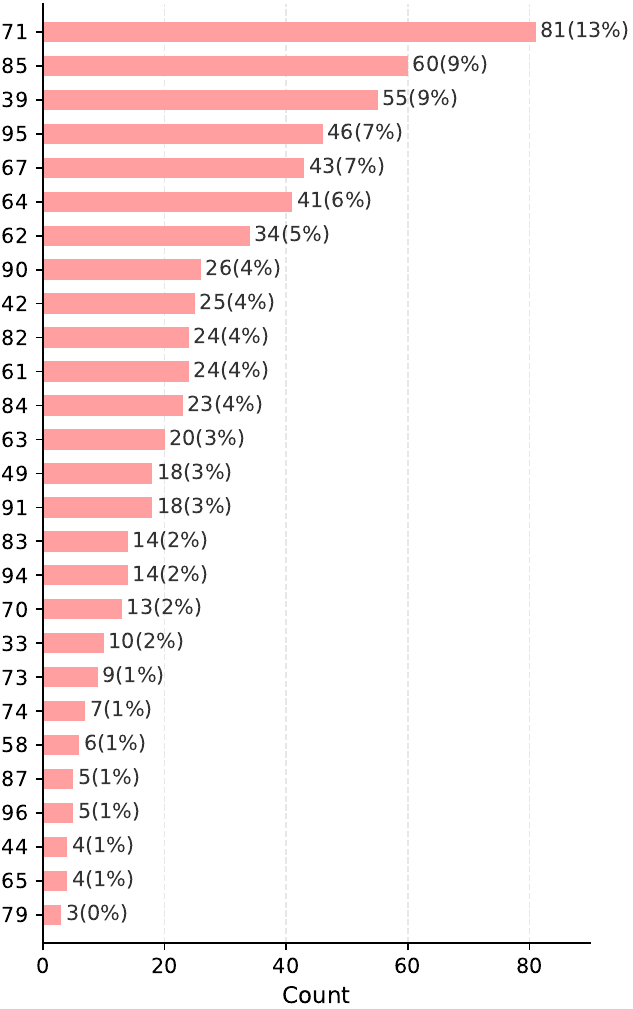}
    \caption{HSCodes distribution on chapter}
  \end{subfigure}
  \caption{Distributions of the first-level product category and HSCode chapter categories.}
  \label{fig:lv1_hs2}
  \vspace{-15pt}
\end{figure}

Figure~\ref{fig:lv1_hs2} presents the distributions of first-level product categories (left) and HSCode chapter categories (right), which closely mirror real-world product distributions. This alignment confirms that \textsc{HSCodeComp} accurately reflects practical international trade scenarios, ensuring that model performance evaluations reliably generalize to real-world applications.

\section{Semantic Distribution of Hierarchical Tariff Rules}
\begin{figure}[H]
    \centering
    \includegraphics[width=0.7\linewidth]{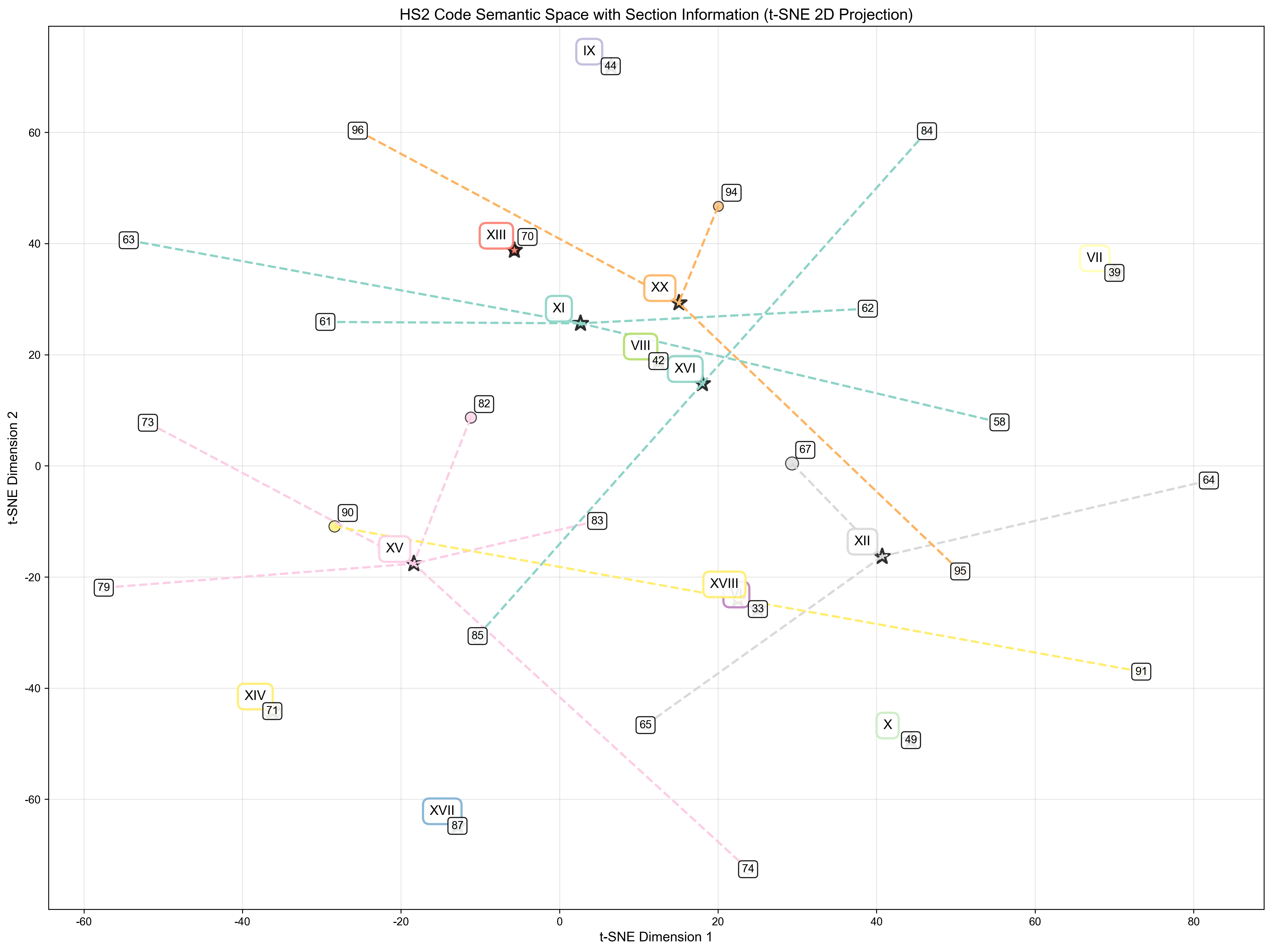}
    \caption{The semantic map of HS chapter titles and notes.}
    \label{fig:hs2_section_semantic_map}
\end{figure}

To assess whether the HSCode taxonomy exhibits clear semantic separation, we generate embeddings of the official English titles and notes for all HS chapters and sections using a sentence embedding model\footnote{\url{https://huggingface.co/sentence-transformers/all-MiniLM-L6-v2}}. We then apply t-SNE to project these embeddings into two dimensions for visualization.
As shown in Figure~\ref{fig:hs2_section_semantic_map}, each point represents a chapter, while each star marks a section's centroid. The visualization reveals significant semantic overlap between adjacent sections: numerous chapters appear closer to neighboring section centroids than to their own section's centroid, and section centroids themselves form overlapping clusters rather than distinct groupings. This pattern indicates that the semantic structure of hierarchical tariff rules lacks clear boundaries—adjacent sections frequently share similar vocabulary and concepts (\textit{e.g.,} distinctions between raw materials and finished goods, or between component parts and complete articles).

\section{More Detailed Experiments on Open-source Agents with Different Backbone LLMs}
\label{appendix.smolagents_different_models}

To investigate how the backbone LLMs affect the performance of the agent system, we conduct more detailed ablation study on four open-source agent systems, replacing the original GPT-5 backbone with Gemini 2.5 Pro.
The experimental results in Table~\ref{tab:agent_backbone_ablation_new} indicate that GPT-5 achieves consistently better performance than the advanced Gemini 2.5 pro model on these open-source agent systems.

\begin{table}[htbp]
    \centering
\begin{tabular}{@{}p{3.5cm}C{1.5cm}C{1.5cm}C{1.5cm}C{1.5cm}C{1.5cm}@{}}
\toprule
\multicolumn{1}{c}{\multirow{2}{*}{Backbone LLM}} & \multicolumn{5}{c}{HSCode Prediction Accuracy} \\ \cmidrule(l){2-6}
\multicolumn{1}{c}{} & 2-digit & 4-digit & 6-digit & 8-digit & 10-digit \\ 
\midrule\multicolumn{6}{c}{SmolAgent}\\\midrule
GPT-5& \textbf{82.28} & \textbf{70.89} & \textbf{59.81} & \textbf{49.05} & \textbf{42.72} \\
Gemini-2.5-Pro& 82.19 & 69.48 & 57.87 & 44.04 & 34.49 \\
Claude 4 Sonnet& 80.69 & 67.09 & 54.11 & 42.25 & 33.70 \\
Qwen-MAX& 77.34 & 63.23 & 42.47 & 26.62 & 17.43 \\
\midrule\multicolumn{6}{c}{Aworld}\\\midrule
GPT-5&\textbf{82.28}&\textbf{70.41}&\textbf{59.18}&\textbf{48.58}&\textbf{41.30}\\
Gemini 2.5 Pro&79.55&66.97&54.70&38.79&29.24\\
\midrule\multicolumn{6}{c}{WebSailor}\\\midrule
GPT-5&\textbf{81.64}&\textbf{70.56}&\textbf{57.27}&\textbf{43.98}&\textbf{35.44}\\
Gemini 2.5 Pro&78.79&67.58&56.21&42.27&31.21\\
\midrule\multicolumn{6}{c}{AgentOrchestra}\\\midrule
GPT-5&82.12&\textbf{70.73}&\textbf{60.44}&\textbf{47.78}&\textbf{41.30}\\
Gemini 2.5 Pro&\textbf{82.27}&69.39&56.97&41.36&30.61\\
\bottomrule
    \end{tabular}
    \caption{The ablation study of backbone models in the open-source agent system.\label{tab:agent_backbone_ablation_new}}
\end{table}

\section{Improvement Gain from Agents}
\begin{figure}[H]
    \centering
    \includegraphics[width=\linewidth]{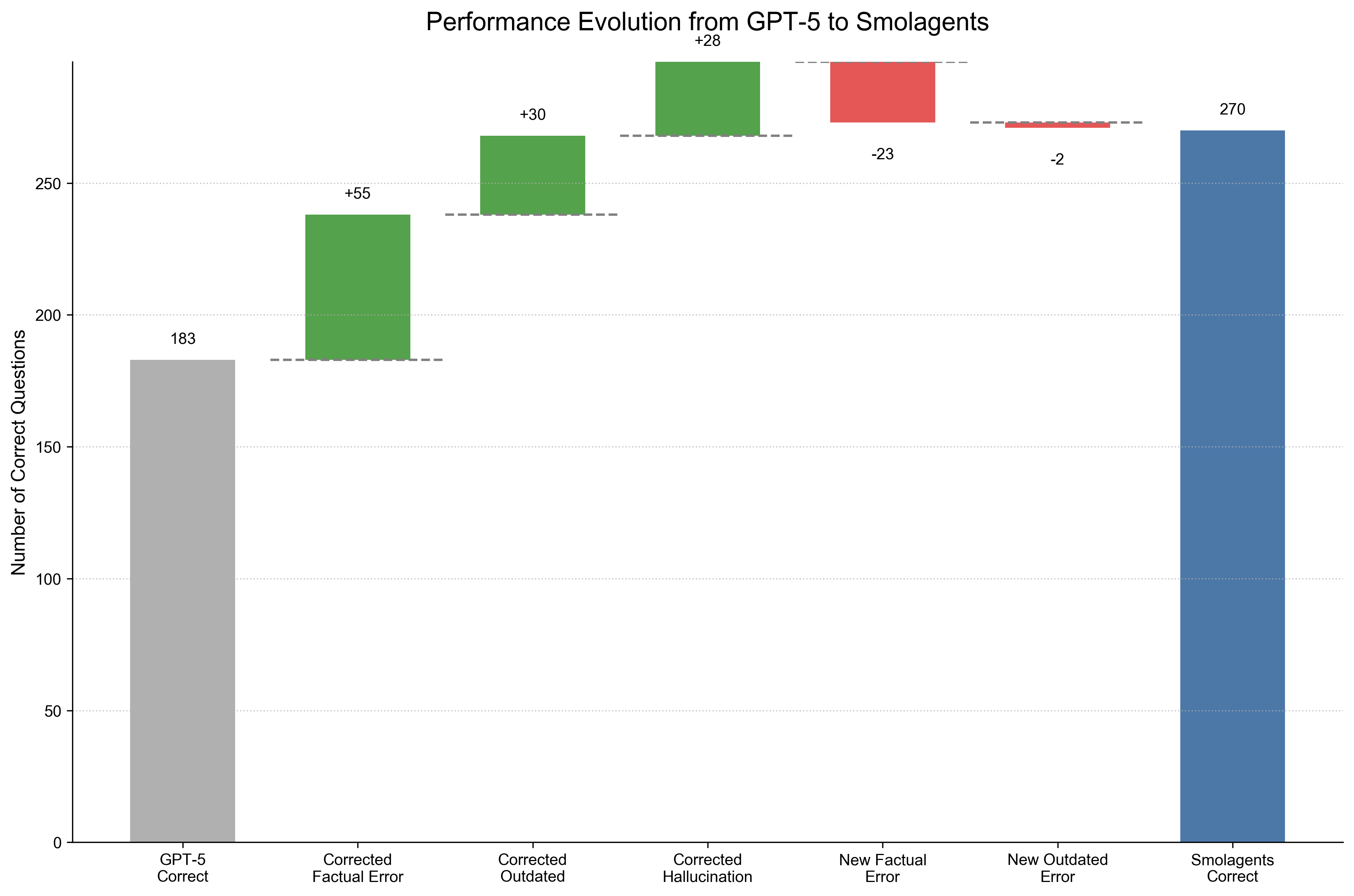}
    \caption{Details of performance gain and loss comparing GPT-5 and Smolagents.}
    \label{fig:agent_llm_improvement}
\end{figure}

\begin{figure}[H]
    \centering
    \includegraphics[width=\linewidth]{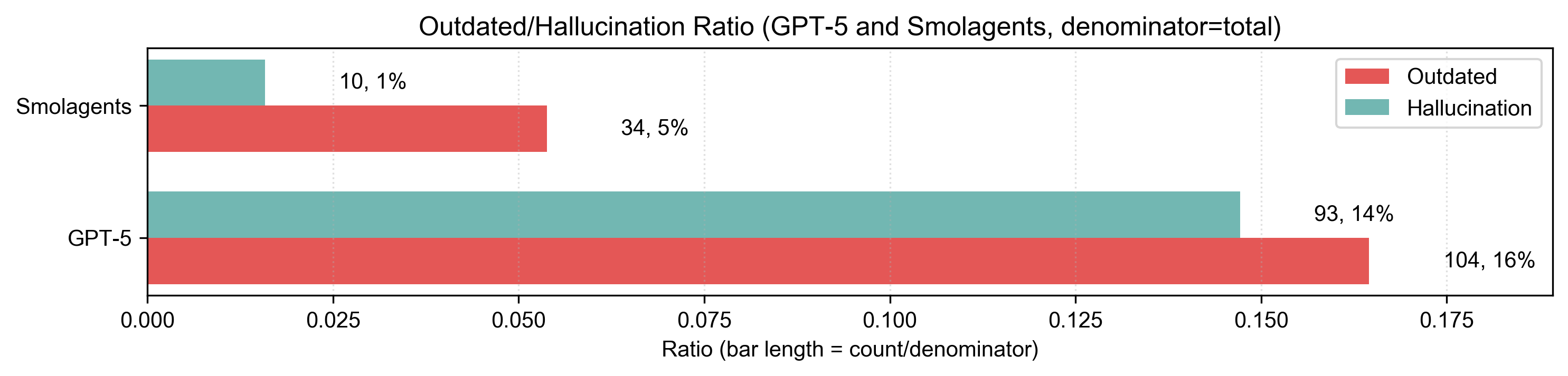}
    \caption{Outdated or hallucination ratio happended on GPT-5 and Smolagent. The numbers are number and ratio of the phenomenon.}
    \label{fig:outdated_hallucination_rate_comparison}
\end{figure}

This waterfall in Figure~\ref{fig:agent_llm_improvement} chart reveals that the superior performance of SmolAgent (GPT-5) are primarily from reducing the outdated and hallucination failures, with 56 corrected samples in \textsc{HScodBench}.
Besides, as shown in Figure~\ref{fig:outdated_hallucination_rate_comparison}, it can be found that the rate of outdated and hallucination are significantly reduced in SmolAgent baseline.

\section{Implementing details and Knowledge Forms\label{appendix:implementation_detail}}
\begin{figure}[H]
    \centering
    \includegraphics[width=0.85\linewidth]{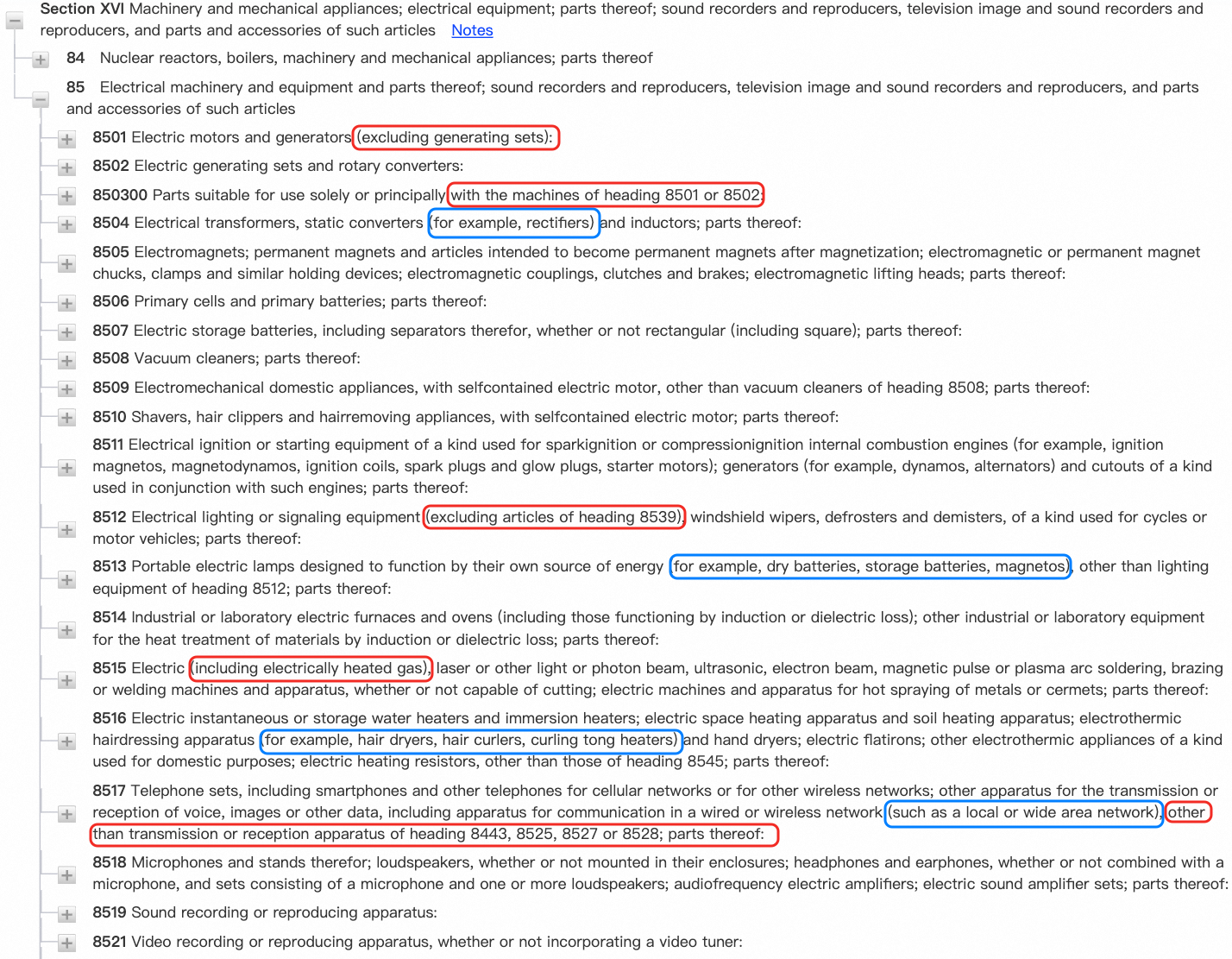}
    \caption{The case of the hierarchical tariff rules.}
    \label{fig:tariff_rules}
\end{figure}

\begin{figure}[H]
    \centering
    \includegraphics[width=0.85\linewidth]{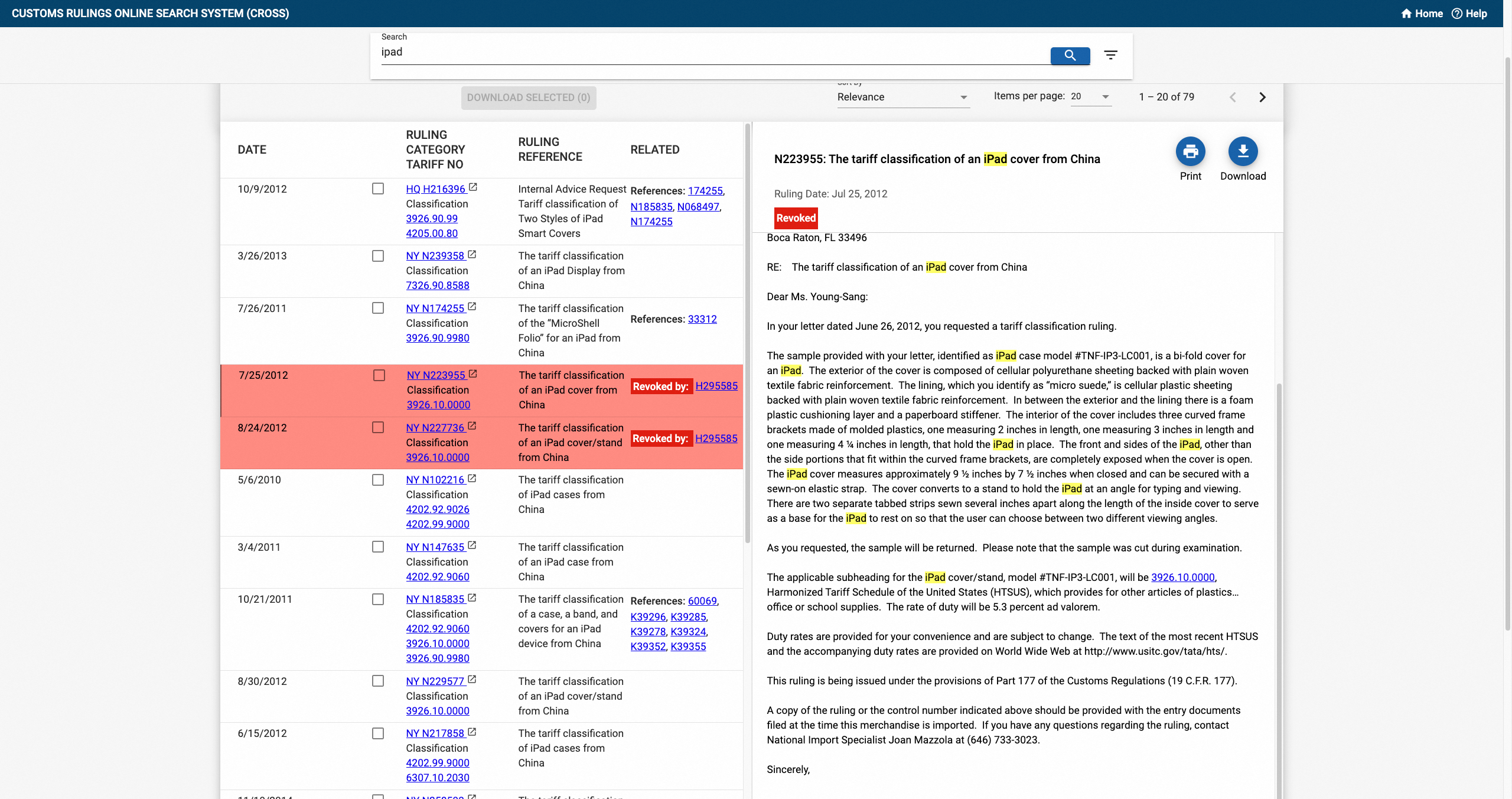}
    \caption{The case of CROSS website that contains the products rulings.}
    \label{fig:cross_example}
\end{figure}

All baseline methods are equipped with search tools to access the CROSS database, hierarchical tariff rules, and other related resources, including human-written knowledge bases and hierarchical decision rules.
The temperature and context window size of LLMs and agents are set to their default configurations.
Moreover, as described in Section~\ref{subsec.main_results}, the hierarchical decision rules, and webpage visit tool are not used during evaluation, since they do not improve the performance of open-source agents. But we do not restrict webpage visit of closed-source agents since we cannot control.
The multi-modal product images are used for open-source agents, i.e. SmolAgents.
The hierarchical decision rules used in our prompts are illustrated in Figure~\ref{prompt_example}.
The hierarchical tariff rules in the eWTP is shown in Figure~\ref{fig:tariff_rules}.
It can be found that the red boxes highlight implicit logical relationships in the tariff rules, such as \textit{excluding articles of heading 8593} and \textit{with the machines of heading 8501 or 8502}. The blue boxes highlight vague descriptions in the tariff rules, such as \textit{``\ldots for example \ldots''} and \textit{``\ldots such as \ldots''}. These cases demonstrate that rule boundaries are ambiguous, posing significant challenges for accurate rule application by the agent.
Moreover, the U.S. Customs Rulings Online Search System (CROSS) interface is shown in Figure~\ref{fig:cross_example}.
As illustrated, the CROSS website contains not only correct precedent results for product HS Codes but also numerous revoked precedents, requiring the agent to carefully evaluate information reliability. Additionally, since the precedent information is presented as plain text emails, the agent must effectively utilize contextual information to perform accurate reasoning.

\begin{figure}[H]
    \centering

\begin{promptbox}[label=prompt_example]

\textbf{The six decision rules for hierarchical tariff rules application}

The following six rules must be applied progressively from Rule 1 to Rule 6, without skipping.

\textbf{Rule 1: Priority of Headings and Notes}

The classification of goods shall be determined primarily according to the terms of the headings (4/6-digit HS codes) and any related Notes.
Subsequent rules shall only be applied if the terms of the headings and the Notes do not suffice for classification.

\textbf{Rule 2: Incomplete/Unfinished Articles and Extension to Materials/Substances}

Rule 2(a): An incomplete or unfinished article (e.g., a bicycle missing wheels), if it has the essential character of the complete article, is to be classified as the complete article.

Rule 2(b): An article consisting of a certain material or substance, which retains its original character after the addition of other materials/substances (e.g., a plastic cup with a metal base), is to be classified according to the original material.

Example:
An unassembled computer motherboard (which already has the function of a motherboard) is classified under heading 8473 (parts of computers).

\textbf{Rule 3: Decision Logic for Goods Classifiable Under Multiple Headings}

When goods are classifiable under two or more headings, classification shall be effected as follows, in order of priority:

Specificity (The more specific description shall be preferred to a more general description);
Essential Character (Determined by the main material, function, or use of the goods);
Last in Numerical Order (If classification cannot be determined otherwise, classify under the heading which occurs last in numerical order).

Example:
An electric toothbrush (which has the attributes of both a "household appliance" and an "oral hygiene tool"):
Specificity: Classified as a "domestic electro-mechanical appliance" (heading 8509) rather than a "toothbrush" (heading 9603).

\textbf{Rule 4: Principle of Closest Analogy}

When goods cannot be classified by applying the preceding three rules, they shall be classified under the heading appropriate to the goods to which they are most similar.

Example:
Imitation leather made from a new material (not listed in the HS) is classified as "artificial leather" (heading 3921).

\textbf{Rule 5: Packing Materials and Containers}

Rule 5(a): Packing materials/containers presented with the goods (e.g., a jewelry box), if normally sold with the goods, are classified with the goods; otherwise, they are classified separately.

Rule 5(b): Reusable packing containers (e.g., metal gas cylinders) are classified separately.

Example:
A glass bottle presented with perfume is classified under the heading for perfume (3303); however, a glass bottle sold separately is classified under 7010.

\textbf{Rule 6: Hierarchical Classification at the Subheading Level}

The classification of goods in the subheadings (6-digit and subsequent HS codes) of a heading shall be determined level by level, first determining the 1-dash subheading (5-6 digits), and then successively the lower-level subheadings. At each level, classification must take into account any Subheading Notes and the relationship between subheadings at the same level.

Example:
After classifying goods under heading 6205 (men's shirts), the subheading is chosen based on material (cotton, man-made fibers, etc.):
620520 (of cotton) or 620530 (of man-made fibres).
\end{promptbox}
\caption{Decision rules defined by human experts.}
\end{figure}

\clearpage
\newpage

\section{Case study of Failure Modes}
We identify six critical failure modes of open-source and closed-source agent systems in \textsc{HSCodeComp}:
(1) \textbf{Premature Decisions}: Agents commit to incorrect classification paths without collecting sufficient evidence (Figure~\ref{fig:early_wrong_search_query} and Table~\ref{tab:closed_source_case_2}-Grok DeepSearch);
(2) \textbf{Information Misprocessing}: Agents overlook or misinterpret key product details, indicating challenges with long-context processing (Table~\ref{tab:closed_source_case_1} and  Figure~\ref{fig:case_noise});
(3) \textbf{Unnecessary Self-Correction}: Agents sometimes predict correct HSCodes initially but revise them incorrectly through excessive critique (Table~\ref{tab:closed_source_case_1}, Gemini Deep Research);
(4) \textbf{Reasoning Hallucination}: Agents generate plausible but factually incorrect reasoning steps (Table~\ref{tab:closed_source_case_1}, Grok DeepSearch);
(5) \textbf{Wrong Rule Application}: Models frequently miss or misuse relevant tariff rules due to their ambiguous descriptions that confuse the reasoning process, resulting in incorrect classification decisions (Figure~\ref{fig:wrong_rule_application});
and (6) \textbf{Lack of Domain Knowledge}: Models exhibit errors due to insufficient domain-specific knowledge, such as misidentify silicone products as rubber instead of plastic (Figure~\ref{fig:lack_domain_knowledge}).
These limitations highlight that \textsc{HSCodeComp} remains challenging for advanced closed-source and open-source systems.

  \begin{center}
    \tcbox[colback=gray!20,colframe=gray!60,boxrule=0.6pt,arc=1mm]{\bfseries Error: Premature Decisions}
  \end{center}

  \vspace{2mm}

  \begin{tcolorbox}[
    title=Task,
    colback=yellow!10,
    colbacktitle=black!80,
    coltitle=white,
    arc=1mm,
    boxrule=0pt,
    breakable]
    \textbf{Product Title}: PEN-F PU Leather Half Case for Olympus PEN-F Digital PENF Camera Brown/Black/Coffee

    \medskip
    \textbf{Product Attributes}:
    \begin{tcolorbox}[colback=yellow!6,colframe=yellow!30,boxrule=0.3pt,arc=0.6mm,sharp corners, left=1.5mm,right=1.5mm,top=1mm,bottom=1mm,breakable]
      \ttfamily\footnotesize
      Origin: Mainland China,\\ Measurement unit: 100000015,\\ Use: Mirrorless System Camera,\\
      Package size - length (cm): 15,\\ Material: PU,\\ Model Number: PEN-F,\\
      Brand Name: NiYi,\\ Type: Camera Bags, Hard Bag,\\ each pack: 1,\\
      Package size - width (cm): 8,\\ Package weight: 0.200,\\ Style: handbags,\\
      Package: No,\\ Package size - height (cm): 5,\\ Sold in: sell by piece
    \end{tcolorbox}

    \textbf{Product Price}: 9.8 USD \\
    \textbf{Category}: Consumer Electronics $\rightarrow$ Accessories \& Parts $\rightarrow$ Camera Bags \& Cases
  \end{tcolorbox}

  \begin{tcolorbox}[title=Search, colback=green!12]
  \textbf{Search query:}
  HTSUS camera case 4202.92 United States 10-digit code camera cases outer surface of plastic sheeting.
  \end{tcolorbox}

\begin{figure}[H] 
  \begin{tcolorbox}[title=Analysis, colback=red!10]

Description of the actual code:

39   Plastics and articles thereof\\
3926 Other articles of plastics and articles of other materials of headings 3901 to 3914:\\
392690 Other:\\
39269099 Other\\
3926909989 Other

  The accurate code is 3926909989. This first search query leads to a wrong direction since it decides the item is under 4202 and ignored it is PU leather which is plastic, and later turns did not realize the ignorance.
  \end{tcolorbox}

\caption{Early wrong search query leads to wrong result.}
\label{fig:early_wrong_search_query}
\end{figure}


  \begin{center}
    \tcbox[colback=gray!20,colframe=gray!60,boxrule=0.6pt,arc=1mm]{\bfseries Error: Real-world noise}
  \end{center}

  \begin{tcolorbox}[
    title=Task,
    colback=yellow!10,
    colbacktitle=black!80,
    coltitle=white,
    arc=1mm,
    boxrule=0pt,
    breakable]
    \textbf{Product Title}: 10/30/60PCS INS Blue Color PVC Sticker \textcolor{red}{Aesthetic} Hand Accounting DIY Decoration \textcolor{red}{Scrapbooking} \textcolor{red}{Korean} \textcolor{red}{Stationery} Supplies 

    \medskip
    \textbf{Product Attributes}:
    \begin{tcolorbox}[colback=yellow!6,colframe=yellow!30,boxrule=0.3pt,arc=0.6mm,sharp corners, left=1.5mm,right=1.5mm,top=1mm,bottom=1mm,breakable]
      \ttfamily\footnotesize
        Origin : Mainland China \\
        Shape : malformed  \\
        Size : M  \\
        Package size - length (cm) : 20  \\
        Material : \textcolor{red}{Plastic}  \\
        Model Number : sticky PVC stickers  \\
        LOT NUM : 1  \\
        Hign-concerned Chemical : None  \\
        PRODUCT UNIT : 100000015  \\
        Brand Name : NoEnName Null  \\
        PACKAGE TYPE : sell by piece  \\
        Package size - width (cm) : 15  \\
        Package weight : 0.030  \\
        Package size - height (cm) : 3  \\
        Age : >3 years old  \\
 
    \end{tcolorbox}

    \textbf{Product Price}: 0.77 USD \hfill \\
    \textbf{Category}: Office \& School Supplies $\rightarrow$ Stationery Sticker $\rightarrow$ Stationery Sticker 
  \end{tcolorbox}

  \begin{tcolorbox}[title=Search query and Prediction, colback=green!12]
  \textbf{Search query:}
  site:rulings.cbp.gov PVC stickers classification 4911.91
  \\
  \textbf{Prediction:}
  \boxed{4911.91.4040} 
 
Classification path 
- Chapter 49: Printed books, newspapers, pictures and other products of the printing industry 
- Heading 4911: Other printed matter, including printed pictures and photographs 
- Subheading 4911.91: Pictures, designs and photographs 
- 8-digit: 4911.91.40: Other (i.e., not lithographs, not posters, and not printed over 20 years prior to import) 
- 10-digit: 4911.91.4040 
 
Decision rationale 
- The product consists of decorative, self-adhesive PVC stickers bearing printed designs (for scrapbooking/decoration). The essential character is the printed picture/design, not the plastic material or adhesive. 

- CBP consistently classifies decorative stickers (including PVC, puffy/3D, epoxy, etc.) as printed pictures under 4911.91 when the printing is the primary purpose, rather than as plastic self-adhesive plates/labels of heading 3919 (which typically applies to pressure-sensitive labels where the printing is merely incidental to labeling). 

- Representative rulings: 

  - NY N300171 (Assorted Stickers from China): classified under 4911.91.4040 as other printed pictures, designs and photographs. 
  
  - NY N262652 (3D PVC cat stickers): classified under 4911.91.4040. 
  
  - NY N273080 (craft stickers, including PVC): classified under 4911.91.4040. 
  
  - NY N301146 (sticker set with PVC printed sheeting): printed picture/design classified in 4911.91.4040. 
  
- The goods are contemporary decorative stickers (not printed over 20 years prior to import), so 4911.91.10.00 does not apply. They are not posters (4911.91.30) and not lithographs on paper (4911.91.20). Therefore, they fall in ``Other,'' 4911.91.40, statistical suffix .4040. 
 
Sources consulted 

- HTSUS 4911.91 structure (USITC search page: 4911.91) and Chapter 49 PDF (2024) 

- CBP Rulings: N300171, N262652, N273080, N301146 (all accessible via rulings.cbp.gov)",

  \end{tcolorbox}

  \begin{tcolorbox}[title=HSCode Description, colback=yellow!12]
  \textbf{Predicted:}\\
49   Printed books, newspapers, pictures and other products of the printing industry; manuscripts, typescripts and plans\\
4911 Other printed matter, including printed pictures and photographs:\\
491191 Pictures, designs and photographs:\\
49119140 Other\\
4911914040 Other

  \textbf{True:}\\
39   Plastics and articles thereof\\
3919 Selfadhesive plates, sheets, film, foil, tape, strip and other flat shapes, of plastics, whether or not in rolls:\\
391910 In rolls of a width not exceeding 20 cm:\\
39191020 Other\\
3919102055 Other
  \end{tcolorbox}

\begin{figure}[H]
  \begin{tcolorbox}[title=Analysis, colback=red!10]
  The accurate code is 3919.10.2055. The material is PVC plastic, and its essential character is defined by its physical form as a self-adhesive sheet in a roll, not the incidental printing on its surface. Therefore, it cannot be classified under Chapter 49 (Printed Matter). The misclassification error stems from \textbf{real-world noise} in the product title. This noise takes the form of subjective marketing terms (e.g., \textbf{Aesthetic}, \textbf{INS}, \textbf{Korean} \textbf{Stationery}) and application-focused words (e.g., \textbf{Decoration}, \textbf{Scrapbooking}). This noise creates a misleading narrative that the printed design is the product's essential character. It directly influences the outcome by shifting focus from the product's objective physical nature (a plastic article) to its subjective use (a decorative image), leading the classifier down the incorrect path of Chapter 49. Instead, as an article of plastic under Chapter 39, the hierarchical rules for self-adhesive plastics in rolls point directly to 3919.10.2055.
  \end{tcolorbox}
  
\caption{Real-world noise forms and analysis.}
\label{fig:case_noise}
\end{figure}


  \begin{center}
    \tcbox[colback=gray!20,colframe=gray!60,boxrule=0.6pt,arc=1mm]{\bfseries Error: Lack domain knowledge}
  \end{center}


  \begin{tcolorbox}[
    title=Task,
    colback=yellow!10,
    colbacktitle=black!80,
    coltitle=white,
    arc=1mm,
    boxrule=0pt,
    breakable]
    \textbf{Product Title}: 20/50/100PCS Pink Nitrile Gloves Disposable Latex Powder Free Gloves for Household Cleaning Beauty Salon Working Kitchen Gloves 

    \medskip
    \textbf{Product Attributes}:
    \begin{tcolorbox}[colback=yellow!6,colframe=yellow!30,boxrule=0.3pt,arc=0.6mm,sharp corners, left=1.5mm,right=1.5mm,top=1mm,bottom=1mm,breakable]
      \ttfamily\footnotesize
Origin : Mainland China \\
Brand Name : INMAICH \\
Type : Disposable \\
each pack : 1 \\
Measurement unit : 100000015 \\
Use : Cleaning \\
Classification : Oversleeve \\
Style : Brief \\
Material : PVC \\
Model Number : Nitrile Disposable Gloves \\
Sold in : sellbypiece 
 
    \end{tcolorbox}

    \textbf{Product Price}: 11.56 USD \hfill \\
    \textbf{Category}: Home \& Garden $\rightarrow$ Household Merchandises $\rightarrow$ Disposable Household Cleaning Products 
  \end{tcolorbox}

  \begin{tcolorbox}[title=Search query and Prediction, colback=green!12]
  \textbf{Search query:}
  HTSUS nitrile disposable gloves HS code 10-digit 4015.19 CBP ruling
  \\
  \textbf{Prediction:}
\boxed{4015.19.1010} 
 
Classification path 
- Section VII: Plastics and Articles Thereof; Rubber and Articles Thereof 
- Chapter 40: Rubber and articles thereof 
- Heading 4015: Articles of apparel and clothing accessories (including gloves), for all purposes, of vulcanized rubber other than hard rubber 
- Subheading 4015.19: Gloves: Other (than surgical) 
- Statistical suffix 4015.19.1010: Seamless, disposable gloves, of vulcanized rubber other than hard rubber 
 
Decision rationale 
- Material and nature: The product is described and marketed as ``nitrile gloves.'' Nitrile is a synthetic rubber (nitrile-butadiene rubber), which places the goods within Chapter 40 (rubber), not plastics. 
- Article type: Gloves for household cleaning/beauty salon/working, i.e., apparel/clothing accessories of rubber → Heading 4015. 
- Use: Non-surgical/industrial/household use → not classifiable as surgical gloves (excludes 4015.11). 
- Construction and disposability: Typical disposable nitrile gloves are dip‑molded and seamless; they are explicitly described as ``Disposable.'' Accordingly, within 4015.19, they fall under the statistical provision for seamless, disposable gloves → 4015.19.1010. 
 
Supporting references 
- CBP NY H86590: classifies disposable nitrile gloves as 4015.19.1010 (seamless, disposable rubber gloves; non‑surgical). 
- CBP NY D83273: disposable nitrile gloves classified under 4015.19.1010. 
 
Notes 
- If the goods were actually made of PVC/vinyl (plastic) rather than nitrile rubber, they would instead fall in Chapter 39 (e.g., 3926.20.xx for plastic gloves). However, based on the stated product title/model (``Nitrile Disposable Gloves''), the correct US HTSUS 10-digit code is 4015.19.1010.
    
  \end{tcolorbox}

  \begin{tcolorbox}[title=HSCode Description, colback=yellow!12]
  \textbf{Predicted:}\\
40   Rubber and articles thereof\\
4015 Articles of apparel and clothing accessories (including gloves, mittens and mitts), for all purposes, of vulcanized rubber other than hard rubber:\\
- Gloves, mittens and mitts:\\
401519 Other:\\
40151911 Seamless\\
4015191110 Disposable\\
4015191150 Other\\
4015195100 Other\\
  \textbf{True:}\\

39   Plastics and articles thereof\\
3926 Other articles of plastics and articles of other materials of headings 3901 to 3914:\\
392620 Articles of apparel and clothing accessories (including gloves, mittens and mitts):\\
39262010 Seamless\\
3926201020 Disposable
  \end{tcolorbox}

\begin{figure}[H]
  \begin{tcolorbox}[title=Analysis, colback=red!10]
This failure is a classic example of \textbf{\textcolor{red}{lacking critical domain knowledge}}, demonstrated by an inability to resolve \textbf{contradictory product data} and the use of an \textbf{outdated HTS code}. The analysis was fundamentally flawed because it prioritized the marketing-oriented \textit{Product Title (Nitrile Gloves)} over the explicit technical attribute, \textit{Material: PVC}. This \textbf{critical error} led the analyst to misclassify the item as a \textbf{rubber} product under \textbf{Chapter 40}. A domain expert would recognize that a \textbf{specific material attribute is a more reliable basis} for classification than a potentially inaccurate title, correctly identifying the item as a \textbf{plastic} article of \textbf{Chapter 39}. Compounding this mistake, the predicted code, \texttt{4015.19.1010}, is an \textbf{\textcolor{red}{obsolete tariff number}}, a definitive sign that the analyst's \textbf{domain knowledge is not current} with the official HTSUS schedule.

  \end{tcolorbox}
  
\caption{Lack of domain knowledge}
\label{fig:lack_domain_knowledge}
\end{figure}


  \begin{center}
    \tcbox[colback=gray!20,colframe=gray!60,boxrule=0.6pt,arc=1mm]{\bfseries Error: Wrong rule application}
  \end{center}


  \begin{tcolorbox}[
    title=Task,
    colback=yellow!10,
    colbacktitle=black!80,
    coltitle=white,
    arc=1mm,
    boxrule=0pt,
    breakable]
    \textbf{Product Title}: Fashion TYPE 2 \textcolor{red}{DIABETIC} Bracelet Soft Silicone Bracelet With Type 2 Diabetes Medical Warning Bracelet Hot Sale 

    \medskip
    \textbf{Product Attributes}:
    \begin{tcolorbox}[colback=yellow!6,colframe=yellow!30,boxrule=0.3pt,arc=0.6mm,sharp corners, left=1.5mm,right=1.5mm,top=1mm,bottom=1mm,breakable]
      \ttfamily\footnotesize
Origin : Mainland China\\
Shape pattern : Round\\
Measurement unit : 100000015\\
Package size - length (cm) : 15\\
Fine or Fashion : Fashion\\
Material : Silicone\\
Gender : Unisex\\
CN : Fujian\\
Item Type : BRACELETS\\
Model Number : SL--268\\
Metals Type : None\\
Hign-concerned Chemical : None\\
Brand Name : NPKDS\\
Function : None\\
each pack : 1\\
Package size - width (cm) : 15\\
Package weight : 0.013\\
Style : TRENDY\\
Setting Type : None\\
Clasp Type : None\\
Compatibility : All Compatible\\
Package size - height (cm) : 2\\
Sold in : sell by piece\\
Bracelets Type : Hologram Bracelets\\
Chain Type : Rope Chain 
 
    \end{tcolorbox}

    \textbf{Product Price}: 1.71 USD \hfill \\
    \textbf{Category}: Jewelry \& Accessories $\rightarrow$ Fashion Jewelry $\rightarrow$ Bracelets
  \end{tcolorbox}

  \begin{tcolorbox}[title=Search query and Prediction, colback=green!12]
  \textbf{Search query:}
  US HTS 7117.90 silicone bracelet imitation jewelry HS code medical alert bracelet
  \\
  \textbf{Prediction:}
  \boxed{7117.90.7500}

Classification path
- Section XIV: Natural or cultured pearls; precious or semiprecious stones; precious metals; imitation jewelry; coin
- Chapter 71: Natural or cultured pearls, precious or semiprecious stones, precious metals, etc.
- Heading 7117: Imitation jewelry
- Subheading 7117.90: Other
- Statistical breakout 7117.90.7500: Valued over 20 cents per dozen pieces or parts; other; of plastics

Decision rationale
- Product is a bracelet made of silicone (a plastic), marketed as a fashion/medical warning bracelet. It is not of precious metal or stone and functions as personal adornment/identification, fitting the terms of ``imitation jewelry'' of heading 7117.
- Material: Silicone → aligns with ``of plastics.''
- Value: Listed price is 1.71 USD per piece, which equals 20.52 USD per dozen, clearly ``valued over 20 cents per dozen.''
- CBP precedent supports this classification for silicone or medical ID bracelets when not of precious metal:
  - NY N127618 (silicone wristband): 7117.90.7500 (valued over \$0.20/dozen; of plastics).
  - H015873 (Medical ID bracelet): classifiable under 7117.90.75 when valued over \$0.20/dozen.
  - NY N121391 (silicone wristband): if valued not over \$0.20/dozen → 7117.90.5500; otherwise use 7117.90.7500. Given this product’s price, 7117.90.7500 applies.

Conclusion
- The correct US 10-digit HTS code for this silicone ``Type 2 Diabetes'' fashion/medical warning bracelet is 7117.90.7500.",
    
  \end{tcolorbox}

  \begin{tcolorbox}[title=HSCode Description, colback=yellow!12]
  \textbf{Predicted:}\\
71   Natural or cultured pearls, precious or semiprecious stones, precious metals, metals clad with precious metal, and articles thereof; imitation jewellery; coin\\
7117 Imitation jewelry:\\
711790 Other:\\
7117907500 Of plastics

  \textbf{True:}\\
39   Plastics and articles thereof\\
3926 Other articles of plastics and articles of other materials of headings 3901 to 3914:\\
392690 Other:\\
39269099 Other\\
3926909989 Other
  \end{tcolorbox}

\begin{figure}[H]
  \begin{tcolorbox}[title=Analysis, colback=red!10]
This failure is a classic case of \textbf{Wrong Rule Application}, where a basic rule was incorrectly prioritized over a decisive, higher-level one. The agent mistakenly applied the simple rule of \textbf{classification by apparent form}, seeing a \textit{bracelet} and defaulting to the seemingly specific \textbf{Chapter 71 (Imitation Jewelry)}. However, the correct, overriding rule that should have been applied is the Rule of \textbf{Essential Character}, which embodies the important principle of \textbf{\textcolor{red}{Function Over Form}} in customs classification. In this case, the text \textbf{"TYPE 2 DIABETIC"} was the key that defined its core \textbf{function}, fundamentally changing the product's essence from an ornament to a \textbf{utilitarian medical alert device}. An article with such an overriding utilitarian function is legally excluded from the scope of \textbf{Chapter 71}. Therefore, the correct rule application mandates that the item be classified by its \textcolor{red}{\textbf{function}} and material under \textbf{Chapter 39}, not by its misleading physical form.

  \end{tcolorbox}
  
\caption{Wrong Rule Application.}
\label{fig:wrong_rule_application}
\end{figure}

\clearpage
\newpage
  \begin{tcolorbox}[
    title=Case 1 of Closed-source Agents,
    colback=yellow!10,
    colbacktitle=black!80,
    coltitle=white,
    arc=1mm,
    boxrule=0pt,
    breakable]
    \textbf{Product Title}: Forklift Safety Cage, 36``x36'' inch Heavy Duty Collapsible Forklift Work Platform,1200LBS Capacity with 4 Universal Wheels

    \medskip
    \textbf{Product Attributes}:
    \begin{tcolorbox}[colback=yellow!6,colframe=yellow!30,boxrule=0.3pt,arc=0.6mm,sharp corners, left=1.5mm,right=1.5mm,top=1mm,bottom=1mm,breakable]
      \ttfamily\footnotesize
Origin:US(Origin)\\
Brand Name:NoEnNameNull\\
Installation Method:Assembly

each pack:1

Package size - width (cm):21

Measurement unit:100000015

Package size - length (cm):25

Package weight:0.900

Package size - height (cm):23

Sold in:sell by piece

Hign-concerned Chemical:None

    \end{tcolorbox}

    \textbf{Product Price}: 1765.15 CNY \hfill \\
    \textbf{Category}: Furniture $\rightarrow$ Outdoor Furniture $\rightarrow$ Garden Furniture Sets
  \end{tcolorbox}

\begin{table}[H]
\centering
\renewcommand{\arraystretch}{1.8} 
\resizebox{0.99\linewidth}{!}{
\begin{tabularx}{\linewidth}{| >{\RaggedRight}p{2cm} | >{\RaggedRight}p{3cm}  | >{\RaggedRight}p{3cm} | >{\RaggedRight}p{3cm} | >{\RaggedRight}p{3.4cm} |}
\hline
\rowcolor{gray!20} 
\textbf{Analysis Dimension} & \textbf{SmolAgent} & \textbf{Gemini DeepResearch} & \textbf{Grok DeepSearch} & \textbf{Manus} \\
\hline

\textbf{Final HTSUS Code} & 
8431.20.0000 (Correct) & 
7326.90.8688 (Incorrect) & 
8428.90.0290 (Incorrect) & 
8427.90.0020 (Incorrect) \\
\hline

\textbf{Core Logic Explained} & 
Based on the core principle of HTSUS Section XVI, Note 2, the cage is a \textcolor{red}{\textbf{part}} as it is 'solely or principally for use with' a forklift (heading 8427). Its design, function, and identity are entirely dependent on the forklift. & 
The argument is based on the \textcolor{red}{\textbf{'part vs. accessory' distinction}}. It posits the cage is not an \textcolor{red}{\textbf{'indispensable' part}}, but an optional 'accessory'. Since accessories are precluded from 8431, classification defaults to its constituent material (steel). & 
Characterizes the cage as a \textcolor{red}{\textbf{functional piece of machinery}}. The rationale is that it enables a new function and \textcolor{red}{\textbf{incorrectly compares it to complex attachments}} with their own mechanics (e.g., rotators, clamps). & 
Characterizes the cage as a \textcolor{red}{\textbf{complete 'aerial work platform'}}. The core argument is that its \textcolor{red}{\textbf{'4 universal wheels' constitute a 'mobile base'}} per the legal notes, thus assembling it into a complete vehicle. \\
\hline

\textbf{Key Flaw Analysis} & 
This approach correctly identifies the product's primary use and applies the controlling legal note directly, which is the standard and most reliable method for classification. & 
This is overthinking because the model became \textcolor{red}{\textbf{fixated on a complex, secondary legal nuance}} (part vs. accessory) while \textcolor{red}{\textbf{ignoring the more direct, primary rule}} ('solely or principally for use with'), leading to an unnecessarily complicated and incorrect conclusion. & 
This is an analysis hallucination because the model \textcolor{red}{\textbf{invents characteristics the product lacks}}, effectively \textcolor{red}{\textbf{treating a passive structure as an active machine}}. The entire analysis is built on this fabricated, non-existent product feature. & 
The model misses key product information by \textcolor{red}{\textbf{misunderstanding the function of a key feature}} (the wheels). It correctly identifies the wheels but misses their trivial context (ground convenience), instead \textcolor{red}{\textbf{mistaking them for a vehicle's chassis}}, which invalidates the entire classification. \\
\hline

\textbf{Failure Modes} & 
\textbf{Correct}  & 
\textbf{Unnecessary Self-Correction}  & 
\textbf{Reasoning Hallucination} & 
\textbf{Information Misprocessing} \\
\hline

\end{tabularx}}
\caption{Comparative Analysis of AI Model Classifications for a Forklift Safety Cage mentioned above.\label{tab:closed_source_case_1}}
\end{table}

  \begin{tcolorbox}[
    title=Case 2 of Closed-source Agents (Cylinder Shock Absorber),
    colback=yellow!10,
    colbacktitle=black!80,
    coltitle=white,
    arc=1mm,
    boxrule=0pt,
    breakable]
    \textbf{Product Title}: ZJSHUYI HR60/80 Pneumatic Cylinder Shock Absorber Adjustable Industrial Used for Manufacturing Equipment

    \medskip
    \textbf{Product Attributes}:
    \begin{tcolorbox}[colback=yellow!6,colframe=yellow!30,boxrule=0.3pt,arc=0.6mm,sharp corners, left=1.5mm,right=1.5mm,top=1mm,bottom=1mm,breakable]
      \ttfamily\footnotesize
Origin:Mainland China\\
product name:Hydraulic buffer\\
Measurement unit:100000015\\
Package size - length (cm):35\\
working environment:Suitable for automated machinery\\
Hign-concerned Chemical:None\\
Certification:CE\\
Brand Name:NoEnName Null\\
product material:iron\\
each pack:1\\
Package size - width (cm):8\\
Working Temperature:-10$^\circ C$~+80$^\circ C$\\
Package weight:9.000\\
Features:Reduce vibration and noise, increase output and extend machine life\\
Package size - height (cm):5\\
Sold in:sell by piece\\
Operating temperature:-10$^\circ C$~+80$^\circ C$
 
    \end{tcolorbox}

    \textbf{Product Price}: 16.73 USD \hfill \\
    \textbf{Category}: Industrial \& Business $\rightarrow$ Industrial Spare Parts $\rightarrow$ Industrial Hardware
  \end{tcolorbox}

\begin{table}[H]
\centering
\renewcommand{\arraystretch}{1.8} 
\resizebox{0.99\linewidth}{!}{
\begin{tabularx}{\linewidth}{| >{\RaggedRight}p{0.1\linewidth} | >{\RaggedRight}p{0.2\linewidth}  | >{\RaggedRight}p{0.2\linewidth} | >{\RaggedRight}p{0.2\linewidth} | >{\RaggedRight}p{0.17\linewidth} |}
\hline
\rowcolor{gray!20} 
\textbf{Analysis Dimension} & \textbf{SmolAgent} & \textbf{Gemini DeepResearch} & \textbf{Grok DeepSearch} & \textbf{Manus} \\
\hline

\textbf{Final HTSUS Code} & 
8487.90.0080 (Correct) & 
8412.21.0075 (Incorrect) & 
8302.49.6085 (Incorrect) & 
8412.31.0080 (Incorrect) \\
\hline

\textbf{Core Logic Explained} & 
Based on the hierarchical structure of HTSUS Chapter 84, the shock absorber is a \textcolor{red}{\textbf{generic machinery part}}. After systematically eliminating more specific headings, it correctly classifies the item in the \textcolor{red}{\textbf{residual heading 84.87}} for parts "not elsewhere specified." & 
Characterizes the product as an \textcolor{red}{\textbf{active hydraulic motor}}. The logic is that because it is a linear-acting hydraulic device, it must be a "motor" under heading 84.12, which is an apparatus that \textcolor{red}{\textbf{generates force or motion}}. & 
It correctly identifies the passive function but then classifies it as a \textcolor{red}{\textbf{simple base metal fitting}}. The argument is that since it's not a motor, its classification defaults to a general heading for \textcolor{red}{\textbf{common hardware and accessories}}. & 
Characterizes the product as an \textcolor{red}{\textbf{active pneumatic motor}}. The rationale is based on the keyword "Pneumatic Cylinder" in the title, concluding it must be an \textcolor{red}{\textbf{actuator that performs work}} under heading 84.12. \\
\hline

\textbf{Key Flaw Analysis} & 
This approach correctly identifies the product's non-specific nature and applies the HTSUS's hierarchical structure and residual headings, which is the standard and most reliable method for such goods. & 
This is a fundamental misunderstanding of the product's function, as the model \textcolor{red}{\textbf{mistakes a passive energy-dissipating device (a damper) for an active power-generating device (a motor)}}. It confuses braking with accelerating. & 
The model makes a decision with insufficient information about HTSUS structure because it \textcolor{red}{\textbf{fails to consider the critical distinction between Chapter 83 (simple fittings) and Chapter 84 (machinery)}} and thus \textcolor{red}{\textbf{underestimates the product's nature}} as a piece of machinery. & 
This is a fundamental misunderstanding of the product's function, as the model is \textcolor{red}{\textbf{misled by an inaccurate keyword}} and \textcolor{red}{\textbf{mistakes a passive damper for an active motor}}, ignoring contradictory product attributes. \\
\hline

\textbf{Failure Modes} & 
\textbf{Correct} & 
\textbf{Information Misprocessing} & 
\textbf{Premature Decisions} & 
\textbf{Information Misprocessing} \\
\hline

\end{tabularx}}
\caption{Comparative Analysis of AI Model Classifications for a Cylinder Shock Absorber mentioned above.\label{tab:closed_source_case_2}}
\end{table}

\section{Case Study of Overthinking}
\label{appendix:overthink}
This section provides an in-depth case studies of open-source and closed-source agent systems, addressing two key observations: (1) the decreased performance brought by overthinking of open-source agents; and (2) the relatively lower performance of state-of-the-art closed-source agents compared to their open-source counterparts.

\subsection{Open-source Agent Cases}
\label{appendix:open_source_failures}

  \begin{center}
    \tcbox[colback=gray!20,colframe=gray!60,boxrule=0.6pt,arc=1mm]{\bfseries Case 1: WebSailor Overthink.}
  \end{center}

  \begin{tcolorbox}[
    title=Task,
    colback=yellow!10,
    colbacktitle=black!80,
    coltitle=white,
    arc=1mm,
    boxrule=0pt,
    breakable]
    \textbf{Product Title}: Men's and Women's Fashion Printed T-shirts, Men's and Women's Fashion Shirts, WWE John Scene,  Winning This Day , Summer, New

    \medskip
    \textbf{Product Attributes}:
    \begin{tcolorbox}[colback=yellow!6,colframe=yellow!30,boxrule=0.3pt,arc=0.6mm,sharp corners, left=1.5mm,right=1.5mm,top=1mm,bottom=1mm,breakable]
      \ttfamily\footnotesize
Origin : Mainland China\\
Package size - length (cm) : 10\\
Gender : Men\\
Fabric Type : Broadcloth\\
Hign-concerned Chemical : None\\
Special Features : Breathable,Lightweight\\
Applicable Scene : Daily\\
Place Of Origin : China (mainland)\\
each pack : 1\\
Applicable Season : Spring And Summer\\
Style : Casual\\
Sleeve Length(cm) : short (4-16inch)\\
Package size - height (cm) : 10\\
Collar : O-Neck\\
Measurement unit : 100000015\\
Tops Type : TEES\\
Material : POLYESTER\\
Item Type : tops\\
Craft of Weaving : Knit\\
CN : Jiangxi\\
Brand Name : NoEnNameNull\\
Package size - width (cm) : 10\\
Product Care Instructions : Machine Wash\\
Hooded : No\\
Pattern Type : 3D\\
Package weight : 0.200\\
Closure Type : Pull On\\
Sold in : sellbypiece
 
    \end{tcolorbox}

    \textbf{Product Price}: 74.96 CNY \hfill \\
    \textbf{Category}: Men's Clothing $\rightarrow$ Tops \& Tees $\rightarrow$ T-Shirts
  \end{tcolorbox}

  \begin{tcolorbox}[title=First step, colback=green!12]
  \textbf{Websailor:}
  
  We need the precise 10-digit HTSUS classification for a men’s knit T-shirt of man-made fiber (polyester). Heading 6109 covers T-shirts, singlets and other vests, knitted or crocheted. Subheading splits by material: 6109.10 (cotton) vs 6109.90 (other fibers). Our item is polyester, so 6109.90. Within 6109.90, the US splits by fiber (man-made fibers vs other), and by sex (men’s/boys vs women’s/girls), often also by ``underwear'' vs ``other.'' The product is an outerwear short-sleeve printed T-shirt (not underwear), for men. The common 10-digit code used by importers and CBP rulings for men’s/boys’ T-shirts of man-made fibers, other than underwear, is 6109.90.1050 (mens or boys). I will confirm using CBP rulings and classification references via web search.
  
name :  search ,  arguments : { query : [ HTSUS 6109.90.1050 men t-shirt} man-made fibers ,  rulings.cbp.gov 6109.90.1050 ,  CBP ruling men polyester knit T-shirt HTSUS ,  HTS 6109.90 man-made fiber T-shirts men boys underwear other ,  HTSUS code men polyester T-shirt 10-digit ]
  \\
  \textbf{Smolagents:}
search query:

HTSUS 6109.90 T-shirts of man-made fibers 10-digit statistical suffix men

  \end{tcolorbox}

  \begin{tcolorbox}[title=HSCode Description, colback=yellow!12]
  \textbf{Websailor wrong prediction:}\\
61   Articles of apparel and clothing accessories, knitted or crocheted
6109 Tshirts, singlets, tank tops and similar garments, knitted or crocheted:
610990 Of other textile materials:
61099010 Of manmade fibers
6109901050 Women's (639)

  \textbf{Smolagents correct prediction:}\\
Section XI Textiles and textile articles Notes
61   Articles of apparel and clothing accessories, knitted or crocheted
6109 Tshirts, singlets, tank tops and similar garments, knitted or crocheted:
610990 Of other textile materials:
61099010 Of manmade fibers
6109901007 Men's (638)
  \end{tcolorbox}

  \begin{center}
    \tcbox[colback=gray!20,colframe=gray!60,boxrule=0.6pt,arc=1mm]{\bfseries Case 2: WebSailor Overthink and directly answer. }
  \end{center}

  \begin{tcolorbox}[title=Task,colback=yellow!10,colbacktitle=black!80,coltitle=white,arc=1mm,boxrule=0pt,breakable]
    \textbf{Product Title}: SPRAY GUN with adapter 400cc Mix tank 1.3/1.8mm nozzle air spray gun with quick-connect wash-free tank paint Mixing Cup sprayer

    \medskip
    \textbf{Product Attributes}:
    \begin{tcolorbox}[colback=yellow!6,colframe=yellow!30,boxrule=0.3pt,arc=0.6mm,sharp corners, left=1.5mm,right=1.5mm,top=1mm,bottom=1mm,breakable]
      \ttfamily\footnotesize
Origin : Mainland China \\
Power Type : Pneumatic \\
Measurement unit : 100000015 \\
Battery Included : No \\
Use : Commercial Manufacture \\
Cup Volume : 400cc \\
Package size - length (cm) : 30 \\
Nozzle Diameter : 1.3mm \\
Model Number : GTI \\
Hign-concerned Chemical : None \\
Feed Type : Gravity \\
Certification : CE \\
Brand Name : NTOOLS \\
Type : HVLP \\
each pack : 1 \\
Package size - width (cm) : 22 \\
Package weight : 1.300 \\
No-Load Speed : 1200rpm \\
Application : Paint Spray Gun \\
Package size - height (cm) : 20 \\
Sold in : sellbypiece

    \end{tcolorbox}

    \textbf{Product Price}: 98.0 USD \hfill \\
    \textbf{Category}: Tools $\rightarrow$ Power Tools $\rightarrow$ Spray Gun
  \end{tcolorbox}

  \begin{tcolorbox}[title=First step, colback=green!12]
  \textbf{Websailor:}

We need the precise 10-digit US HS (HTSUS) code for a pneumatic paint spray gun with a cup (HVLP gravity-feed). These are generally classifiable under Chapter 84, heading 8424, which covers mechanical appliances for projecting or spraying liquids or powders. Specifically, subheading 8424.20 covers ``Spray guns and similar appliances.'' The 10-digit breakouts in the HTSUS usually distinguish paint spray guns from other spray guns (e.g., agricultural, steam/sand blasting, etc.). Based on industry practice and CBP rulings, paint spray guns (including HVLP air spray guns with attached cups) fall in 8424.20.10 at the 8-digit level, with a final statistical suffix commonly .90 for ``Other'' paint spray guns (not parts). Therefore, the full 10-digit code is 8424.20.1000 or 8424.20.1090 depending on the structure. Recalling the U.S. HTSUS, the statistical suffix for ``Spray guns and similar appliances: Spray guns: Paint spray guns'' is 8424.20.1000 (often the sole 10-digit). To ensure accuracy, we align with typical import classifications for HVLP automotive paint spray guns with gravity cups, which are consistently entered under 8424.20.1000. No electrical drive; compressed air powered; used for paint spraying—squarely within that provision. No need to use other headings like 8467 (tools) because 8424 is specific to spraying devices. Now present classification path and rationale and the final 10-digit code in LaTeX box.
\boxed{8424.20.1000} \ldots
  \textbf{SmolAgents:} \\
  search query: HTS 842420 paint spray guns 10 digit statistical suffix\\
  \end{tcolorbox}

  \begin{tcolorbox}[title=HSCode Description, colback=yellow!12]
  \textbf{Websailor wrong prediction:}\\
84   Nuclear reactors, boilers, machinery and mechanical appliances; parts thereof
8424 Mechanical appliances (whether or not hand operated) for projecting, dispersing or spraying liquids or powders; fire extinguishers, whether or not charged; spray guns and similar appliances; steam or sand blasting machines and similar jet projecting machines; parts thereof:
842420 Spray guns and similar appliances:
8424201000 Simple piston pump sprays and powder bellows

  \textbf{Smolagents correct prediction:}\\
84   Nuclear reactors, boilers, machinery and mechanical appliances; parts thereof
8424 Mechanical appliances (whether or not hand operated) for projecting, dispersing or spraying liquids or powders; fire extinguishers, whether or not charged; spray guns and similar appliances; steam or sand blasting machines and similar jet projecting machines; parts thereof:
842420 Spray guns and similar appliances:
8424209000 Other
  \end{tcolorbox}


\section{Case Study of Multi-modal Information in Agent Systems}
\label{appendix:multimodal_agent_ablation}
This section presents experimental case studies comparing SmolAgent (GPT-5) performance with and without product image processing.







\begin{table*}[h]
\centering
\small
\renewcommand{\arraystretch}{1} 
\setlength{\tabcolsep}{4pt}      

\begin{tabularx}{\textwidth}{@{} >{\centering\arraybackslash}p{0.12\textwidth} X X X X @{}}
\toprule
\textbf{Image} &
\textbf{Product Information} &
\textbf{Non-vision HS (10-digit) + brief rationale} &
\textbf{With-vision HS (10-digit) + brief rationale} &
\textbf{Vision advantage (reason category)} \\
\midrule

\includegraphics[width=\linewidth]{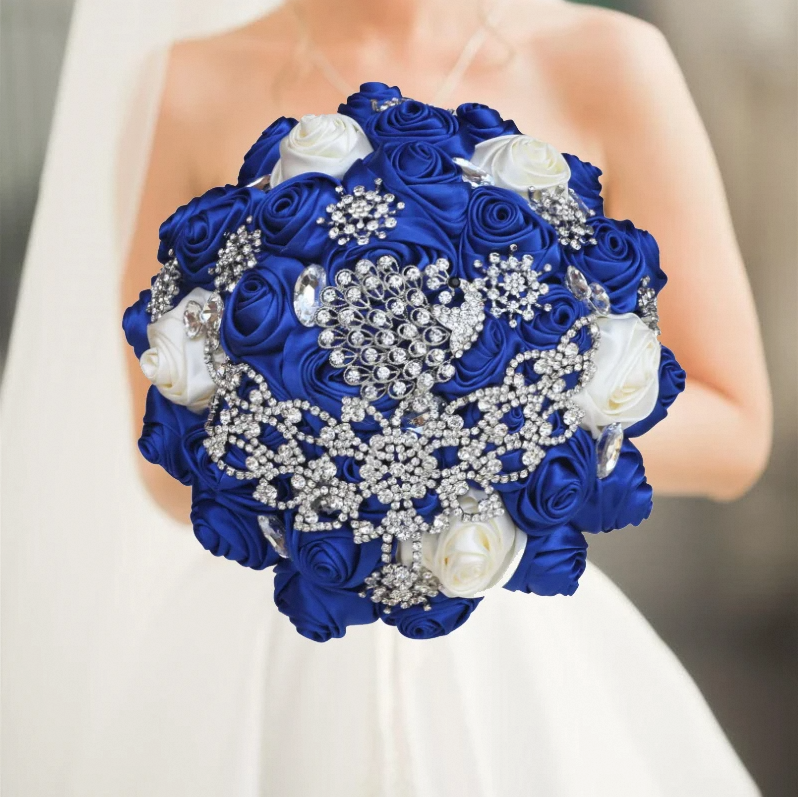} & \vspace{-55pt}
\textbf{Title:} Royal Wedding Bouquet Rhinestone Bride and Bridesmaid Hand Flowers Handmade Bridal Bouquet.
\par \textbf{Category:} Artificial Decorations.
\par \textbf{Attributes:} Material: Silk (natural textile), not plastics; bouquet of artificial flowers with rhinestones. & \vspace{-55pt}
\textbf{6702.90.3500} (Artificial flowers of \emph{man-made fibers}).
\par Assumes ``silk flowers'' = polyester; treats bouquet as MMF artificial flowers. & \vspace{-55pt}
\textbf{6702.90.6500} (Artificial flowers, \emph{other than man-made fibers}).
\par Images/text indicate silk fabric (natural fiber); rhinestones do not change essential character (GRI 3(b)). & \vspace{-55pt}
Material fiber identification (natural silk vs assumed polyester). \\
\midrule

\includegraphics[width=\linewidth]{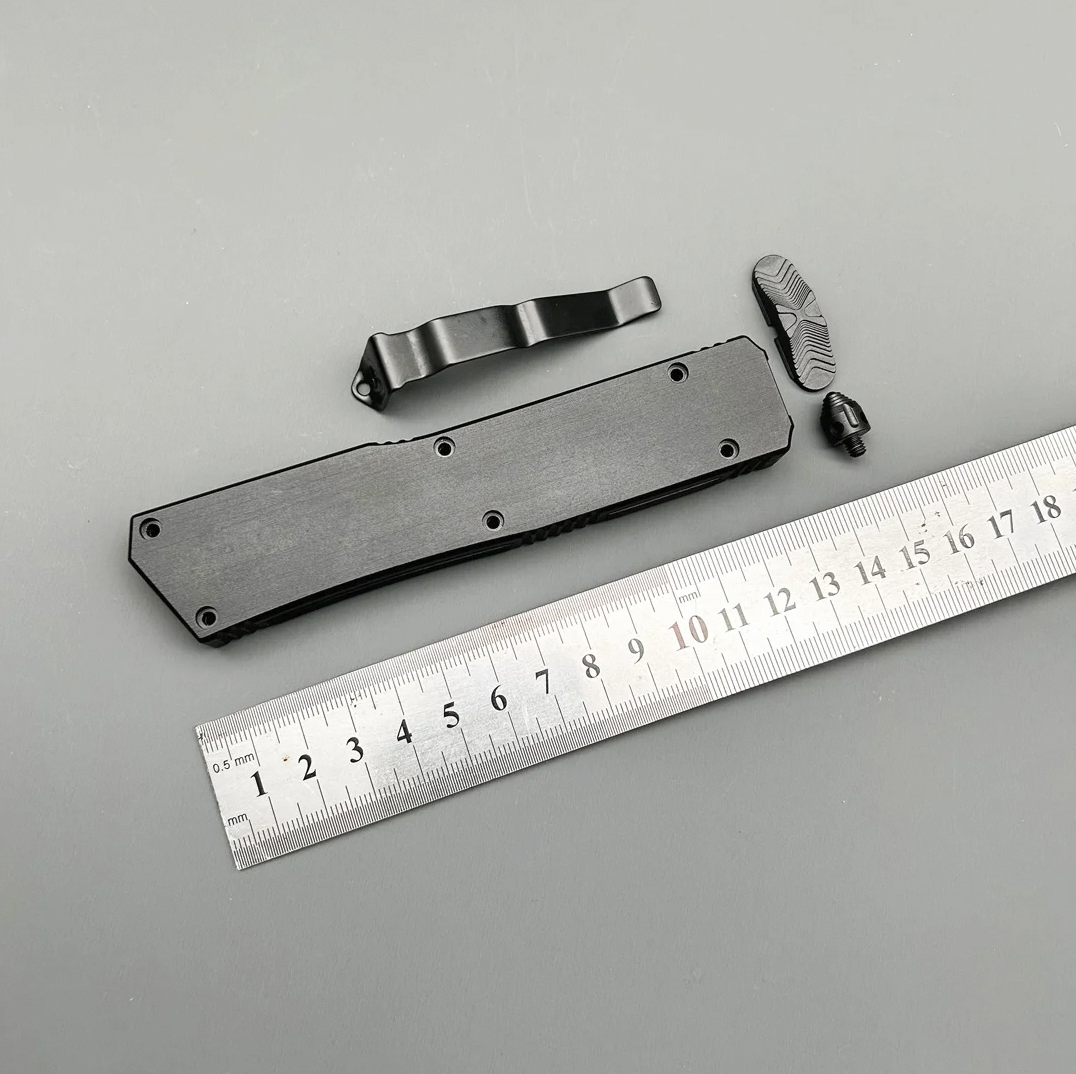} & \vspace{-55pt}
\textbf{Title:} OTF knife parts, aviation aluminum tactical handle kit.
\par \textbf{Category:} Hand Tools / Knife accessories.
\par \textbf{Attributes:} Handle parts only, no blade present; aluminum body with clip, actuator, screws. & \vspace{-55pt}
\textbf{8211.93.0035} (Folding/pocket knives).
\par Interprets listing as a \emph{complete folding knife} rather than components. & \vspace{-55pt}
\textbf{8211.95.9000} (Handles of base metal, other).
\par Visuals confirm no blade; essential character is the base‑metal handle assembly (parts provision applies). & \vspace{-55pt}
Completeness parts identification (parts‑only vs whole article). \\
\midrule

\includegraphics[width=\linewidth]{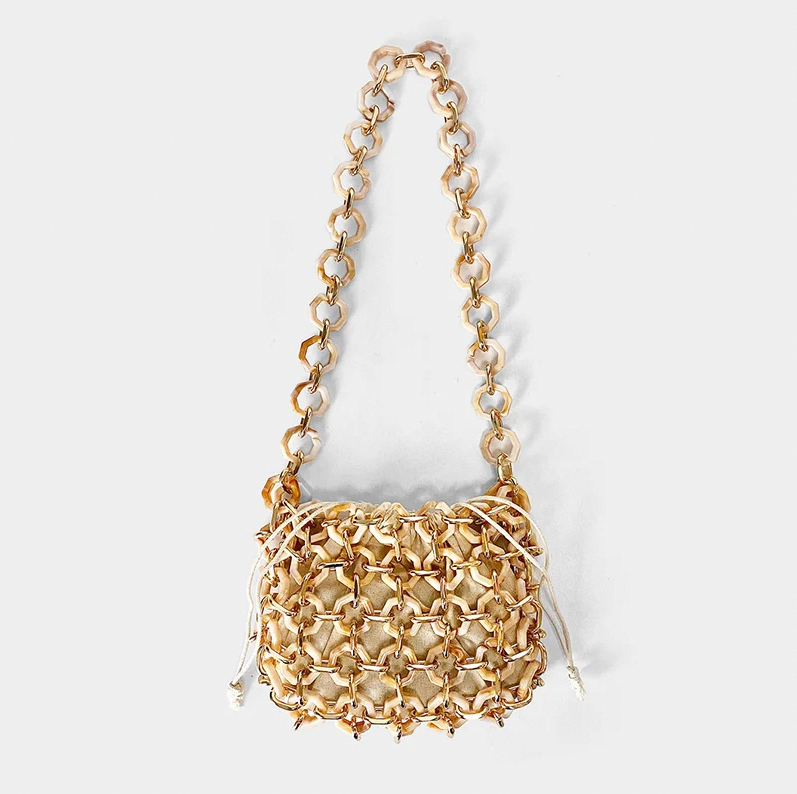} & \vspace{-55pt}
\textbf{Title:} Acrylic buckle, beaded/openwork shoulder bag with inner pouch.
\par \textbf{Category:} Women's Handbags.
\par \textbf{Attributes:} Outer surface is beads openwork lattice (ABS acrylic) , 
not plastic sheeting; linen pouch is interior. & \vspace{-55pt}
\textbf{4202.22.1500} (Handbags with outer surface of \emph{sheeting of plastics}).
\par Assumes exterior is plastic sheeting based on ``ABS.'' & \vspace{-55pt}
\textbf{4202.29.9000} (Other).
\par Images show beads rings openwork, not ``sheeting of plastics''; classification by outer surface (Additional U.S. Note 2 to Ch. 42). & \vspace{-55pt}
Structural outer‑surface feature (beads openwork vs plastic sheeting). \\
\midrule

\includegraphics[width=\linewidth]{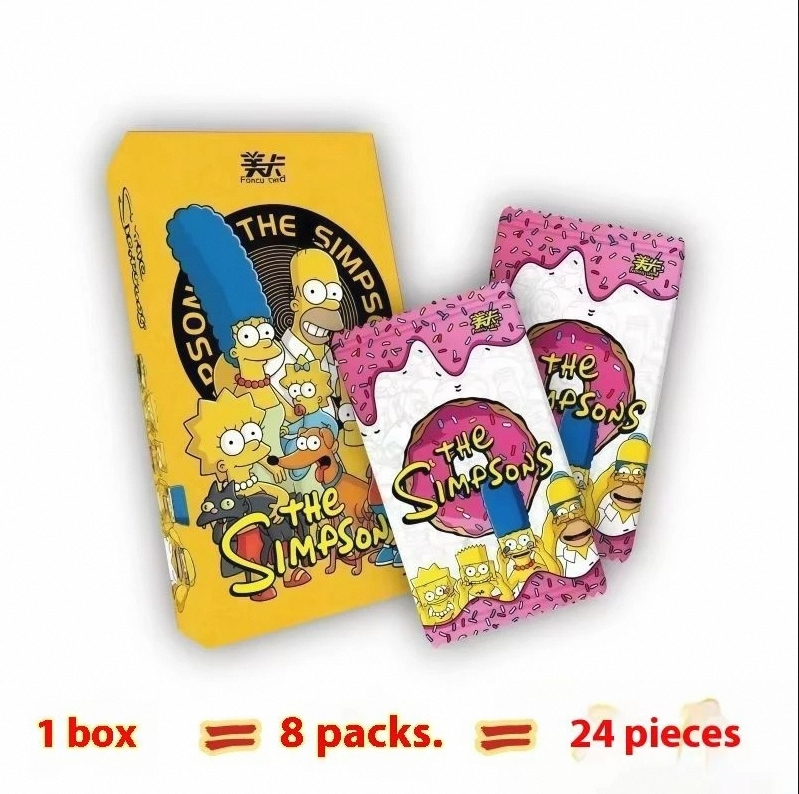} & \vspace{-55pt}
\textbf{Title:} The Simpsons character collectible paper cards.
\par \textbf{Category:} Printed matter / collectibles.
\par \textbf{Attributes:} Cards bear pictures only; no rules-based deck specified; no lithographic process evidence. & \vspace{-55pt}
\textbf{4911.99.6000} (Other printed matter, often linked to lithographic printing).
\par Assumes lithographic process without explicit evidence. & \vspace{-55pt}
\textbf{4911.91.4040} (Pictures, designs and photographs; other).
\par Visuals confirm picture-only cards and lack of lithographic evidence/criteria; not a playing-card game. & \vspace{-55pt}
Process evidence-based exclusion (no lithography evidence; pictures-only). \\
\bottomrule
\end{tabularx}
\caption{Vision vs Non-vision: Four Representative Cases with Key Evidence and Reasons.\label{tab:vision_vs_nonvision}}
\end{table*}

\end{document}